\documentclass[twoside,11pt]{article}
\usepackage{jmlr2eNoEditor}
\usepackage{amsmath}
\usepackage{amsfonts}
\usepackage{verbatim}
\usepackage[usenames]{color}
\usepackage{graphicx}
\usepackage{stmaryrd}
\usepackage[mathscr]{eucal}
\usepackage{ifthen}
\input xy
\xyoption{all}
\newboolean{showProofs}
\newtheorem{clam}{Claim}
\newcommand{\p}[0]{p^{\phantom{g}}}

\newcommand{\ranglel}[0]{\rangle^{\phantom{j}}}

\newcommand{\qed}{\hfill\BlackBox}
\newcommand{\defn}[1]{(def \ref{#1})}
\setboolean{showProofs}{true}

\ShortHeadings{Towards a Sound  Theory of Adaptation for the SGA}{Burjorjee}
\firstpageno{1}

\begin{document}

\title{Towards a Sound Theory of Adaptation for the \\Simple Genetic Algorithm}

\author{\name Keki Burjorjee \email kekib@cs.brandeis.edu   \\
       \addr
       Computer Science Department\\
       Brandeis University\\
       Waltham, MA 02454-9110, USA}

\editor{}

\maketitle

\begin{abstract}
The pace of progress in the fields of  Evolutionary Computation and Machine Learning is currently  limited --- in the former field, by the improbability of making advantageous extensions to evolutionary algorithms when their capacity for adaptation is poorly understood, and in the latter by the difficulty of finding effective semi-principled reductions of \emph{hard} real-world problems to relatively \emph{simple} optimization problems. In this paper we explain why a theory which can accurately explain the simple genetic algorithm's remarkable capacity for adaptation has the potential to address both these limitations. We describe what we believe to be  the impediments --- historic and analytic --- to the discovery of such a theory and highlight the negative role that the building block hypothesis (BBH) has played. We argue based on experimental results that a fundamental limitation which is widely believed to constrain the SGA's adaptive ability (and is strongly implied by the BBH) is in fact illusionary and does not exist. The SGA therefore turns out to be more powerful than it is currently thought to be. We give conditions under which it becomes feasible to numerically approximate and study the multivariate marginals of the search distribution of an infinite population SGA over multiple generations even when its genomes are long, and explain why this analysis is relevant to the riddle of the SGA's remarkable adaptive abilities.
\end{abstract}


\section{Introduction}\label{intro}

The practice of Machine Learning research can be characterized as the effective semi-principled reduction of learning problems to problems for which robust and  efficient solution techniques exist --- ideally ones with provable bounds on their use of time and space. In a recent paper \citet{journals/jmlr/BennettP06} describe the synergistic relationship between the fields of machine learning (ML) and mathematical programming (MP). They remark:
\begin{quote}
``Optimization lies at the heart of machine learning. Most machine learning problems reduce to optimization problems. Consider the machine learning analyst in action solving a problem for some set of data. The modeler formulates the problem by selecting an appropriate family of models and massages the data into a format amenable to modeling. Then the model is typically trained by solving a core optimization problem that optimizes the variables or parameters of the model with respect to the selected loss function and possibly some regularization function. In the process of model selection and validation, the core optimization problem may be solved many times. The research area of mathematical programming theory intersects with machine learning through these core optimization problems" \citep{journals/jmlr/BennettP06}.\end{quote}

\noindent Later \citeauthor{journals/jmlr/BennettP06}  imply that when the targets of ML reductions have been optimization problems, they have for the most part been the convex optimization problems within the MP pantheon.

\begin{quote}"Convexity plays a key role in mathematical programming. Convex programs minimize convex optimization functions subject to convex constraints ensuring that every local minimum is always a global  minimum. In general, convex problems are much more tractable algorithmically and theoretically. The complexity of nonconvex problems can grow enormously. General nonconvex programs are NP-hard" \citep{journals/jmlr/BennettP06}.\end{quote}

The close relationship between ML and MP arguably exists because MP provides ML with a set of crisp, well-defined problems along with algorithmic solvers  that come with guarantees on their use of time and space. To state this using metaphors from software engineering, the well-defined convex optimization problems are interfaces that MP publishes, and the provably efficient and robust algorithmic solvers of MP implement these interfaces.

Let us differentiate, in this paper, between optimization and adaptation. We define \emph{optimization} as the procurement of one or more points of optimal or close-to-optimal value, and \emph{adaptation} as the generation of points of increasing value over time. Given this definition, to say that the target problems of Machine Learning reductions are optimization problems is to fudge the truth somewhat. While the Mathematical Programming community indeed seems to be almost completely concerned with the procurement of optimal or close to optimal points, ML researchers aren't interested in optimization per se but in the means by which it is achieved in most MP algorithms, i.e. adaptation. In fact optimization is often prevented in machine learning algorithms --- using a ``technique" named \emph{early-stopping} --- to prevent overfitting. In other words, robust, efficient adaptation is the modus operandi of most convex optimization algorithms, and for the most part, it is this modus operandi that makes these algorithms interesting to Machine Learning researchers.

The interface-problems published by the MP community give ML researchers useful targets to hit; if a ML researcher works out a semi-principled reduction of a class of learning problems to one of MP's interface-problems, there are off-the-shelf algorithms within MP which allow her to quickly test whether her reduction is effective.

Because of the emphasis that the ML community place on guarantees of robustness and efficiency. when the targets of ML reductions have been optimization problems, they have for the most part, been restricted to being convex optimization problems within MP. These problems are rather simple as adaptation problems go  --- every local optimum is also a global optimum, or stated differently, there are no local optima. Rather heroic feats of ingenuity are therefore necessary in order to obtain effective semi-principled reductions of hard problems to these simple optimization problems. The difficulty of obtaining such reductions is currently a fundamental limitation on the pace of progress within ML.

The SGA \citep{Mitchell:1996:IGA} is an adaptation algorithm which mimics natural sexual evolution. It has been \emph{directly} applied to a large number of hard real-world problems and has often succeeded in generating solutions of remarkably high-quality. To be sure, some amount of thought is required to ``massage'' these problems into a form which allows the SGA to operate successfully on them (e.g. choices must be made about the fitness function used and the way solutions are encoded as bitstrings), but unlike the case in machine learning this massaging is largely ad-hoc, an outcome more of trial and error than principled reasoning. The resulting problems are almost certainly hard ones (non-convex),  with objective functions that are riddled with local optima. It is a testament to the adaptive power of the SGA that it nevertheless often produces solutions of remarkably high quality. Given these successes one might expect a great deal of interest in SGAs from the machine learning community. That this is not the case speaks to an unfortunate shortcoming of GA research. There is no dearth of one-off problems that  SGAs have adequately solved. However GA researchers have yet to publish a single \emph{class }of problems such that  a) SGAs are likely to perform robust, efficient adaptation when applied to problems in this class, and b) the class is likely to be useful as the target of ML reductions. For the sake of brevity we will loosely define such a class of problems as an SGA-Easy/ML-Useful class. We believe that when such problem classes are found the ML community will begin to take a greater interest in GA research. The future relationship between the GA and ML communities might then be similar to the one that currently exists between MP and ML.

As mentioned above SGAs commonly adapt high-quality solutions to problems which are almost certainly contain large numbers of local optima. It is reasonable therefore to suspect that there exists an SGA-Easy/ML-Useful class of hard non-convex  problems and that the identification of this class will significantly ease the burden of obtaining novel ML reductions. We believe that the identification of such a problem class will go hand in hand with the discovery of a theory which can give a satisfying explanation of the adaptive capacity of the SGA. Such a theory does not currently exist.

\subsection{The Dubious History of the Building Block Hypothesis} \label{checkeredHistory}
Perceptions of the abilities and limitations of the SGA (and hence the kinds of problems that it can and cannot solve)  have been heavily influenced by a theory of adaptation called the \emph{building block hypothesis} \citep{Goldberg:1989:GAS, Mitchell:1996:IGA, holland75:_adapt_natur_artif_system, journals/ec/Holland00}. This theory of adaptation has its genesis in the following idea: maybe small groups of closely located co-adaptive alleles propagate within an evolving population of genomes in much the same way that single adaptive alleles do in Fisher's theories of sexual evolution \citep{Fisher58}. Holland called such groups of alleles \emph{building blocks}. This  idea can be taken one step further: maybe small groups of co-adaptive \emph{building blocks} propagate within an evolving population of genomes in much the same way that single building blocks do. Such groups can be thought of as higher-level building blocks. Pursuing this idea to the fullest extent, maybe co-adaptive groups of higher-level building blocks propagate in much the same way as ordinary building blocks do to yield building blocks of an even higher level, and so on and so forth in hierarchical fashion with the building-blocks of higher levels  being comprised of co-adaptive groups of lower-level building blocks. Let us call this this idea \emph{hierarchical building block assembly}.

 \citet{holland75:_adapt_natur_artif_system} saw in hierarchical building block assembly a way out from the problem that epistasis \citep{episAndEvolProc} poses for Fisher's theory of sexual evolution. He also believed that hierarchical building block assembly, if implemented efficiently, could serve as a useful problem solving technique. He argued that a genetic algorithm that he called a genetic plan can implement hierarchical building block assembly, and moreover does so efficiently. He offered the genetic plan as a model of natural sexual evolution and also as a useful technique for finding solutions to adaptation problems with non-convex objective functions. The main theoretical tool that he used in his argument has come to be called the schema theorem \citep{Goldberg:1989:GAS, Mitchell:1996:IGA}. However neither the schema theorem, nor any of Holland's other theoretical analyses fully support his claim that simple genetic algorithms are capable of efficiently implementing hierarchical building block assembly . Given the boldness of his claim and the large leaps of intuition that Holland makes in order to support it, the absence of experimental support in \citep{holland75:_adapt_natur_artif_system} is rather conspicuous (even more so given that simple, computationally unintensive, proof-of-concept experiments are not difficult to conceive of.  See, \citealp{mitchell:1992:rrgaflgp}, and \citealp{forrest93relative}) . It would not have been surprising therefore if the genetic plan had been relegated to the history books as an algorithm that did not fulfill its raison d'etre --- to support its inventor's hunch about the utility of hierarchical building block assembly as a theory of adaptation for natural sexual evolutionary systems, and to support its inventor's hunch that hierarchical building block assembly can be efficiently implemented. What seems to have saved the SGA from this fate is the curious matter of its utility.

In the years following the publication of Holland's seminal work \citep{holland75:_adapt_natur_artif_system}, the SGA was successfully used to adapt high-quality solutions to different sorts of real world and toy problems with non-convex objective functions. In an unfortunate twist of reasoning hierarchical building block assembly became the de-facto explanation for the success of the SGA. This explanation came to be called the building block hypothesis. Despite its name, the building block hypothesis was treated more as an assumption than as a hypothesis.  Hierarchical building block assembly had aesthetic appeal, and the building block hypothesis had Holland's unqualified endorsement \citep{Holland92}. Therefore the building block hypothesis was readily accepted by most within the GA community. Some even went so far as to tout the success of SGAs as evidence of the veracity of the building block hypothesis or as evidence that hierarchical building block assembly is a useful search technique for a wide variety of search problems. Consider the following confused passage from one of the first text books on genetic algorithms:\begin{quote}
``...the building block hypothesis has held up in many different problem domains. Smooth, unimodal problems, noisy multimodal problems, and combinatorial optimization problems have all been attacked successfully using virtually the same reproduction-crossover-mutation [S]GA."\citep{Goldberg:1989:GAS}\end{quote}
The early support that the building block hypothesis enjoyed accounts for the deep impact it has had and continues to have on the course of research in genetic algorithms as well as other fields of evolutionary computation such as genetic programming.

Recently the building block hypothesis has been sharply criticized  for lacking adequate theoretical support. The most forceful criticism that we are aware of has been levied by \citet{conf/gecco/WrightVR03}: ``The various claims about [S]GAs that are traditionally made under the name of the \emph{building block hypothesis} have, to date, no basis in theory, and, in some cases, are simply incoherent''.  On the empirical side experimental results have been obtained which straightforwardly cast doubt upon the ability of a simple genetic algorithm to efficiently implement hierarchical building block assembly \citep{mitchell:1992:rrgaflgp,forrest93relative}. In response to these experimental results a silent transition has occurred within the field of genetic algorithms: hierarchical building block assembly has gone from being thought of as the abstract process that SGAs implement to being thought of as a normative process that SGAs \emph{mis}-implement. Even though this transition between intellectual positions is completely specious it is now widely assumed that SGAs work because they manage to ``fudge" hierarchical building block assembly. Many new genetic algorithms have been constructed to compensate for the perceived short-comings of the GA ---e.g. messy GA, \citep{goldberg:1989:mgamafr,Goldberg:1989:GAS,desInnov}, LLGA \citep{harik97learning, desInnov}, CGA \citep{journals/tec/HarikLG99}, ECGA \citep{harik99linkage}, cohort GA \citep{journals/ec/Holland00}, FDA \citep{muehlenbein:1999:fdaadf}, LFDA \citep{muehlenbein:2001:earsd}, BOA \citep{Pelikan:98aa,desInnov}, hBOA \citep{pelikan:2001:HBOABOANLS}, SEAM \citep{oai:eprints.ecs.soton.ac.uk:12006,watsonBook},etc. The inventors of these algorithms claim, or at least imply, that their algorithms are better than the SGA at its own game --- hierarchical building block assembly.  In many circles within the GA community the curious matter of frequent utility of SGAs is now considered closed.

For a case in point of the kind of sleight of hand that we are discussing consider the following: conceding that there is little evidence that SGAs can efficiently and robustly implement hierarchical building block assembly, \citet{journals/ec/Holland00} remarks, ``Are [S]GA's,  then, a robust approach to all problems in which building blocks play a key role? By no means! After years of investigation we still have only limited information about the [S]GA's capabilities for exploiting building blocks". Later he \emph{asserts} that ``the very essence of good GA design is retention of diversity, furthering exploration, while exploiting building blocks already discovered", and presents a new genetic algorithm, the Cohort Genetic Algorithm, and argues that it implements this essence (see \citep{HuafengPEI:2001:GECCO} for evidence that it does not).

The field of genetic algorithms is both a scientific field as well as an engineering domain. Heedful science and meticulous engineering can often work synergistically. However when the boundary between science and engineering begins to blur, dogma and misplaced faith can beleaguer the practice of both, to wit, a system that is useful in practice, but does not implement a hypothetical mechanism may receive reduced attention, whereas the mechanism, far from being dismissed according to the basic norms of science may become the holy grail of the engineering goals of the field.

A theory that explains why a system exhibits a particular behavior can influence perceptions of  how the system can behave, and also of how it cannot. Of the two kinds of perceptions, the latter kind is often judged in retrospect to be the greater impediment to the discovery of a new theory that can explain and predict the behavior of the system with greater accuracy. This is because by influencing perceptions of how the system cannot behave a theory implicitly determines the ``domain of the impossible" and in doing so it steers researchers away from considering certain possibilities. Yet it is precisely amongst these "\emph{im}possibilities" that the seeds of a new more accurate theory often lie.

One of the two goals of this paper is to challenge the widespread belief that the SGA cannot increase the frequency of a low order schema with above-average fitness when the defining length of that schema is high (i.e. when the defining bits of that schema are widely dispersed). This belief can be traced back to Holland's original treatise on genetic algorithms \citep{holland75:_adapt_natur_artif_system} and goes hand in hand with belief in the building block hypothesis (and variations thereof). In section \ref{rescuing} we provide an argument based on experimental evidence that this belief is misplaced. We believe that this errant belief will be judged in retrospect to have been a significant impediment to the discovery of a sound theory of adaptation for the SGA.

\subsection{A Type of Analysis Which Might Yield an Explanation for the SGA's Capacity for Adaptation (And Why This Kind of Analysis is Difficult)} \label{TowPrincTheory}
Our second goal is to present theoretical results which, in our opinion, are likely to advance efforts to discover a theory of adaptation for the SGA.

The \emph{search distribution} of an SGA in some generation is a particular kind of distribution over the set of all genomes. This distribution  is implicitly determined by the bitstrings in the population of the previous generation, their fitness values, the chosen selection scheme, and the variation operators of the SGA. The bitstrings in the current generation can be thought to be generated by monte-carlo sampling  from this search distribution. The search distribution is a very useful concept because it gives one a clean way to conceptually distinguish between the ``deterministic component" and the ``stochastic component" of the effect of selection and variation in each generation. Another way to say this is that the implicit creation of a new search distribution from the individuals in some generation is the only part of the adaptation process over which the selection and variation operators exert any ``deterministic" control. Beyond that, chance is the sole determiner of which individuals are actually generated in the next generation. As the size of the population tends towards infinity the role that chance plays in determining the composition of populations diminishes. In the limit as the population size tends to infinity, chance plays no role at all. In this case the search distribution in some generation exactly describes the composition of the population in that generation. In this case  it is also possible \emph{in principle} to exactly determine the search distribution after any number of generations. We say `in principle' because (as we will soon discuss) this is computationally infeasible in practice when the genome are long.

An infinite population SGA (IPSGA) is a mathematical model of an SGA with a large but finite population. Therefore studies that examine how the search distribution of an IPSGA changes over multiple generations have the potential to shed light on the ``deterministic component" of the multi-generational effect of selection and variation on the search distribution of an SGA with a large but finite population. Such knowledge may lead to a satisfying theory of adaptation for the SGA.

That said, not all studies of effects of evolution on the search distribution of an IPSGA are equally likely to yield theories of adaptation. We believe that studies that examine the effects of evolution over a small number of generations are more likely to be useful than those that examine the effects of evolution in the asymptote of time, namely dynamical systems analyses of the fixed points of IPSGAs (e.g.  \citealp{vose:1999:sgaft}). This is for two reasons. Firstly an analysis of the fixed points of a dynamical system does not reveal how or why the system reaches those fixed points. Adaptation in IPSGAs is a transient, not an asymptotic, phenomenon. Therefore a study of the fixed points of  IPSGAs is unlikely to yield answers about how adaptation occurs. Secondly, recall that an IPSGA is an \emph{inexact} mathematical model of an SGA with a large but finite population. The longer the timescale under consideration, the more likely it is that the search distribution of the IPSGA will  diverge from that of the SGA --- even one with a large population.

One promising way to study the changes in the search distribution of an IPSGA over multiple generations is to study the multivariate marginals of this changing distribution. For an arbitrary IPSGA this becomes computationally infeasible very quickly as the length of the bitstring genomes increases (for reasons we will soon explain). In this paper we will derive conditions under which such a study is feasible for genomes of arbitrary length. Stating the above in the language of schemata \citep{holland75:_adapt_natur_artif_system,Goldberg:1989:GAS,Mitchell:1996:IGA}, we will derive conditions under which it becomes computationally feasible to track the frequencies of schemata in a schema family \citep{conf/gecco/WrightVR03} over multiple generations even when the the bitstring genomes of the IPSGA are long.

Let us briefly review why calculating the frequencies of an IPSGA over multiple generations is computationally infeasible for long genomes. An IPSGA with genomes of length $\ell$ can be modeled by a set of $2^\ell$ coupled difference equations. For each genome in the search space there is a corresponding state variable which gives the frequency of the genome in the population, and a corresponding difference equation which describes how the value of that state variable in some generation can be calculated from the values of the state variables in the previous generation. A naive way to calculate the frequency of some schema over multiple generations is to numerically iterate the IPSGA over many generations, and in each generation, to sum the frequencies of all the genomes that belong to the schema.  The simulation of one generation of an IPSGA with a genome set of size $N$ has time complexity $O(N^3)$, and an IPSGA with bitstring genomes of length $\ell$ has a genome set of size $N=2^\ell$. Hence, the time complexity for a  numeric simulation of one generation of an IPSGA is $O(8^\ell)$ . See \citep[p.36]{vose:1999:sgaft} for a description of how the Fast Walsh Transform  can be used to bring this bound down to $O(3^\ell)$.) Even when the Fast Walsh Transform is used, computation time still increases exponentially with $\ell$. Therefore for large $\ell$ the naive way of calculating the frequencies of schemata over multiple generations clearly becomes computationally infeasible\footnote{Vose reported in 1999 that computational concerns force numeric simulation to be limited to cases where $\ell\leq20$}.

Holland's schema theorem \citep{holland75:_adapt_natur_artif_system,Goldberg:1989:GAS,Mitchell:1996:IGA} was the first theoretical result which allowed one to calculate (albeit imprecisely) the frequencies of schemata after a single generation. The crossover and mutation operators of a GA can be thought to destroy some schemata and construct others. Holland only considered the destructive effects of these operators. His theorem was therefore an inequality. Later work \citep{conf/icga/StephensW97} contained a theoretical result which gives exact values for the schema frequencies after a single generation. Unfortunately for IPSGAs with long bitstrings this result does not straightforwardly suggest conditions under which schema frequencies can be numerically calculated over multiple generations in a computationally feasible way.

\subsection{The Promise of Coarse-Graining}
Coarse-graining is a technique that has widely been used to study aggregate properties (e.g. temperature) of many-body systems with very large numbers of state variables (e.g. gases). This technique allows one to reduce some system of difference or differential equations with many state variables (called the fine-grained system) to a  new system of difference or differential equations that describes the time-evolution of a smaller set of state variables (the coarse-grained system). The state variables of the fine-grained system are called the microscopic variables and those of the coarse-grained system are called the macroscopic variables. The reduction is done using a surjective non-injective function between the microscopic state space and the macroscopic state space. This function is called the partition function. States in the microscopic state space that share some key property (e.g. energy) are projected to a single state in the macroscopic state space. The reduction is therefore `lossy', i.e. information about the original system is typically lost. Metaphorically speaking, just as a stationary light bulb projects the \emph{shadow} of some moving 3D object onto a flat 2D wall, the partition function projects the changing state of the fine-grained system onto states in the state space of the coarse-grained system.

The term `coarse-graining' has been used in the Evolutionary Computation literature to describe different sorts of reductions of the equations of an IPSGA. Therefore we now clarify the sense in which we use this term. In this paper a reduction of a system of equations must satisfy three conditions to be called a coarse-graining. Firstly, the number of macroscopic variables should be smaller than the number of microscopic variables. Secondly, the new system of equations must be completely self-contained in the sense that the state-variables in the new system of equations must \emph{not} be dependent on the microscopic variables. Thirdly,  the dynamics of the new system of equations must `shadow' the dynamics described by the original system of equations in the sense that if the projected state of the original system at time $t=0$ is equal to the state of the new system at time $t=0$ then at any other time $t$, the projected state of the original system should be closely approximated by the state of the new system. If the approximation is instead an equality then the reduction is said to be an \emph{exact} coarse-graining. Most coarse-grainings are not exact. This specification of  coarse-graining is consistent with the way this term is typically used in the scientific literature. It is also similar to the definition of coarse-graining given in \citep{journals/tcs/RoweVW06} (the one difference being that in our specification a coarse-graining is assumed not to be exact unless otherwise stated).

Suppose the vector of state variables $\mathbf x^{(t)}$ is the state of some system at time $t$ and the vector of state variables $\mathbf y^{(t)}$ is the state of a coarse-grained system at time $t$.  Now, if the partition function projects $\mathbf x^{(0)}$ to $\mathbf y^{(0)}$, then, since none of the state variables of the original system are needed to express the dynamics of the coarse-grained system, one can determine how the state of the coarse-grained system $\mathbf y^{(t)}$ (the shadow state) changes over time without needing to determine how the state in the fine-grained system $\mathbf x^{(t)}$ (the shadowed state) changes. Thus, even though for any $t$, one might not be able to determine $\mathbf x^{(t)}$, one can always be confident that $\mathbf y^{(t)}$ is its projection. Therefore, if the number of state variables of the coarse-grained system is small enough, one can numerically iterate the dynamics of the (shadow) state vector $\mathbf y^{(t)}$ without needing to determine the dynamics of the (shadowed) state vector $\mathbf x^{(t)}$.

In this paper we give sufficient conditions under which it is possible to coarse-grain the dynamics  of an IPSGA such that the macroscopic variables are the frequencies of the  family of schemata in some schema partition. If the size of this family is small then, regardless of the length of the genome, one can use the coarse-graining result to numerically calculate the approximate frequencies of these schemata over multiple generations in a computationally tractable way. Given some population of bitstring genomes, the set of frequencies of a family of schemata describe the multivariate marginal distribution of the population over the defined loci of the schemata. Thus another way to state our contribution is that we give sufficient conditions under which the multivariate marginal distribution of an evolving population over a small number of loci can be numerically approximated over multiple generations regardless of the length of the genomes.

We stress that our use of the term `coarse-graining' differs from the way this term has been used in other publications within the evolutionary computation literature. For instance in \citep{Stephens:2003:gecco} the term `coarse-graining' is used to describe a reduction of the IPSGA equations such that each equation in the new system is similar in form to the equations in the original system. However the state variables in the new system are defined in terms of the state variables in the original system. Therefore a numerical iteration of the the new system is only computationally tractable when the length of the genomes is relatively short. Elsewhere the term coarse-graining has been inconsistently used to refer to ``a collection of subsets of the search space that covers the search space''\citep{conf/gecco/ContrerasRS03}, and as ``just a function from a genotype set to some other set"\citep{interPopConstaints}.

\subsubsection{A Previous Coarse-Graining Result}


\citet{conf/gecco/WrightVR03} have shown that the frequency dynamics of the genomes of a non-selective IPSGA  can be coarse-grained such that the macroscopic variables are the frequencies of a family of schemata. However they argue that the dynamics of a regular selecto-mutato-recombinative IPSGA  cannot be similarly coarse-grained ``except in the trivial case where fitness is a constant for each schema in a schema family"\citep{conf/gecco/WrightVR03}. Let us call this condition \emph{schematic fitness invariance}. Wright et. al. imply that it is so severe that it renders the coarse-graining result essentially useless.

This negative result holds true when there is no constraint on the initial population. In this paper we show that if we constrain the class of initial populations then it \emph{is} possible to obtain a similar coarse-graining under a much \emph{weaker} constraint on the fitness function. The constraint on the class of initial populations that is required for our coarse-graining is not onerous; it is easily met by a population that is uniformly distributed over the genome set.

\subsection{Structure of this Paper}

The rest of this paper is organized as follows:  in the next section we define the basic mathematical objects and notation which we use to model the dynamics of an infinite population evolutionary algorithm (IPEA). This framework is very general; we make no commitment to the data-structure of the genomes, the nature of mutation, the nature of recombination,or the number of parents involved in a recombination. We do however require that selection be fitness proportional.  In section \ref{coarsenability}  we define the concepts of semi-coarsenablity, coarsenablity and global coarsenablity which allow us to formalize a useful class of exact coarse-grainings. In section  \ref{variationCoarsenability} and section \ref{selectionCoarsenability} we prove some stepping-stone results about selection and variation. We use these results in section \ref{evolutionCoarsenability} where we prove that an IPEA that satisfies certain abstract conditions can be coarse-grained. The proofs in sections \ref{selectionCoarsenability} and \ref{evolutionCoarsenability} rely on lemmas which have been relegated to and proved in the appendix. As stated above, the theoretical results obtained in sections \ref{variationCoarsenability}--\ref{evolutionCoarsenability} are very general, and are applicable to any IPEA which meets the premises of the theorems in those sections. In sections \ref{AlgebraOfAmbivalence} and \ref{simVarConsts} we develop theoretical machinery which allows us to apply the results of these sections to an IPSGA. In section \ref{suffConds} we specify concrete conditions under which IPSGAs with long genomes and non-trivial fitness functions can be coarse-grained such that the macroscopic variables are the frequencies of a family of schemata and the fidelity of the coarse-graining is likely to be high. Our argument is informal and requires the reader to make small leaps of intuition. In section \ref{expval} we explain why a direct experimental validation of the results of section \ref{suffConds} is computationally infeasible. We then make certain key assumptions and modeling decisions which allows us to indirectly validate these results. In section \ref{rescuing} we use the uncontroversial modeling decision that we make in section \ref{expval} to experimentally show that an SGA is capable of increasing the frequency of a low-order schema with higher than average fitness, even when the defining length of that schema is high. Our experiments show that the widespread belief that an SGA cannot do such a thing is misplaced. In our conclusion we reiterate the importance of obtaining a well-founded theory which explains the adaptive capacity of the SGA and specify the concrete contributions that we have made towards this goal.

\section{Mathematical Preliminaries}
                                                                                                                \label{MathematicalPreliminaries}
                                                                                                                Let $X,Y$ be sets and let $\beta:X\rightarrow Y$ be some function. For any $y\in Y$ we use the notation
                                                                                                                $\langle y\rangle_{\!\beta}^{\phantom g} $ to denote the pre-image of $y$, i.e. the set
                                                                                                                $\{x\in X\,|\,\beta(x)=y\}$. For any subset $A\subseteq X$ we use the notation $\beta(A)$ to denote the set $\{y\in Y|\,\beta(a)=y \text{ and } a\in A\}$

                                                                                                                As in \citep{ToussaintThesis}, for any set $X$ we use the notation $\Lambda^X$ to denote
                                                                                                                the set of all distributions over $X$, i.e. $\Lambda^X$ denotes set $\{f:X\rightarrow
                                                                                                                [0,1] \,\,|\,\, \sum_{x\in X} f(x)=1\}$. For any set $X$, let $0^X:X\rightarrow\{0\}$ be
                                                                                                                the constant zero function over $X$. For any set $X$, an $m$-parent
                                                                                                                transmission function \citep{MontgomerySlatkin05011970,kinnear:altenberg,toussaint:03-foga} over $X$  is an element of the set
                                                                                                                \[\bigg\{T:\prod_1^{m+1} X\rightarrow[0,1]\,\,\bigg|\,\,\forall x_1,
                                                                                                                \ldots,x_m\in  X, \sum_{x\in X}T(x,x_1,\ldots, x_m)=1\bigg\}\]

                                                                                                                Extending the notation introduced above, we denote this set by $\Lambda^X_{m}$. Following \citep{ToussaintThesis}, we use conditional probability notation in our denotation of transmission functions. Thus an $m$-parent transmission function $T(x,x_1,\ldots,x_{m})$ is denoted $T(x|x_{1},\ldots,x_{m})$.

                                                                                                                A transmission function can be used to model the individual-level effect of mutation, which operates on one parent and produces one child, and indeed the individual-level effect of any variation operation which operates on any numbers of parents and produces one child.

                                                                                                                For some genome set $G$, let $T_1\in \Lambda^G_m$ and $T_2 \in \Lambda^G_n$ be
                                                                                                                transmission functions that model $m$ and $n$-parent variation operations respectively.
                                                                                                                Then the effect of applying these two operations one after the other can be modeled by a
                                                                                                                third transmission function: suppose the $m$-parent variation operation is applied to the
                                                                                                                output of the $n$-parent operation, then the transmission function that models the effect
                                                                                                                of the composite variation is given by $T_1\circ T_2$, where composition of transmission
                                                                                                                functions is defined as follows:

                                                                                                                \begin{definition}\,\,\, \textsc{(Composition of Transmission Functions)} For any $T_1\in{\Lambda^X_m}, T_2\in{\Lambda^X_n}$, we define the
                                                                                                                composition of $T_1$ with $T_2$, denoted $T_1\circ T_2$, to be a transmission function in
                                                                                                                $\Lambda^X_n$ given by \begin{align*}(T_1\circ
                                                                                                                T_2)(x|y_1,\ldots,y_n)=\sum_{\substack{(x_1,\ldots,x_m)\in\\\prod_1^m
                                                                                                                X}}T_1(x|x_1,\ldots,x_m)\prod_{i=1}^mT_2(x_i|y_1,\ldots,y_n)\end{align*}
                                                                                                                \end{definition}

                                                                                                                Our scheme for modeling EA dynamics is based on the one used in \citep{ToussaintThesis}. We model the genomic populations of an EA as distributions over the genome set. The population-level effect of the evolutionary operations of an EA is modeled by mathematical operators whose inputs and outputs are such distributions.

                                                                                                                The expectation operator, defined below, is used in the definition of the selection operator, which follows thereafter.

                                                                                                                \begin{definition}\,\,\textsc{(Expectation Operator)}\label{WeightedAverageOperator}
                                                                                                                Let $X$ be some finite set, and let $f:X\rightarrow \mathbb R^+$ be some function. We define the expectation operator $\mathcal E_f:\Lambda^X\cup 0^X \rightarrow \mathbb
                                                                                                                R^+\cup\{0\}$ as follows:\begin{center}
                                                                                                                \[\mathcal E_f(p) =\sum\limits_{x\in X}f(x)p(x)\]\end{center}
                                                                                                                \end{definition}

                                                                                                                The selection operator is parameterized by a fitness function. It models the effect of fitness proportional selection on a population of genomes.
                                                                                                                \begin{definition}\,\,\textsc{(Selection Operator)}\label{SelectionOperator}
                                                                                                                Let $X$ be some finite set and let $f:X\rightarrow \mathbb R^+$ be some function. We
                                                                                                                define the \emph{Selection Operator} $\mathcal S_f:\Lambda^X \rightarrow \Lambda^X$ as
                                                                                                                follows: \[(\mathcal
                                                                                                                S_fp)(x)=\frac{f(x)p(x)}{\mathcal E_f(p)}\]
                                                                                                                \end{definition}

                                                                                                                The population-level effect of variation is modeled by the variation operator. This operator is parameterized by a transmission function which models the effect of variation at the individual level.

                                                                                                                \begin{definition}\,\,\textsc{(Variation Operator\footnote{also called the Mixing Operator
                                                                                                                in \citep{vose:1999:sgaft} and \citep{ToussaintThesis}})}\label{Transmission functionOperator} Let
                                                                                                                $X$ be a countable set, and for any $m\in \mathbb N^+$, let $T\in \Lambda^X_{m}$ be a transmission
                                                                                                                function over $X$. We define the variation operator $\mathcal V^{\phantom |}_T:\Lambda^X
                                                                                                                \rightarrow \Lambda^X$ as follows: \[(\mathcal V^{\phantom |}_Tp)(x) =\sum_{\substack{(x_1,\ldots,x_m)\\\in\,\prod_1^mX}}
                                                                                                                T(x|x_1,\ldots,x_m)\prod_{i=1}^mp(x_i)\]
                                                                                                                \end{definition}

                                                                                                                The next definition describes the projection operator (previously used in
                                                                                                                \citep{vose:1999:sgaft} and \citep{ToussaintThesis}). A projection operator that is parameterized by some function $\beta$ `projects' distributions over the domain of $\beta$, to distributions over its co-domain.

                                                                                                                \begin{definition}\,\,\textsc{(Projection Operator)} \label{ProjectionOperator}
                                                                                                                Let $X$ be a countable set, let $Y$ be some set, and let $\beta:X\rightarrow Y$ be a function. We define the
                                                                                                                projection operator, $\Xi_\beta:\Lambda^X\rightarrow \Lambda^Y$ as follows:
                                                                                                                \[(\Xi_\beta \p)(y)=\sum_{x\in\langle y\rangle_{\!\beta}^{\phantom g}}p(x)\]
                                                                                                                and call $\Xi_\beta \p$ the $\beta$-projection of $p$.
                                                                                                                \end{definition}

                                                                                                                \section{Formalization of a Class of Coarse-Grainings}\label{coarsenability}

                                                                                                                The following definition introduces some convenient function-related terminology.
                                                                                                                \begin{definition}\,\,\textsc{(Theme map, Theme Set, Themes, Theme Class)}\label{ThemeThemeSetThemeClass} Let $X$, $K$ be sets and let
                                                                                                                $\beta:X\rightarrow K$ be a surjective function. We call $\beta$ a \emph{theme map}, call the co-domain $K$ of $\beta$ the \emph{theme
                                                                                                                set} of $\beta$, call any element in $K$ a \emph{theme} of $\beta$, and call the pre-image $\langle k
                                                                                                                \rangle_{\!\beta}^{\phantom g}$ of some $k\in K$, the \emph{theme class} of $k$ under $\beta$.
                                                                                                                \end{definition}

                                                                                                                The next definition formalizes a class of coarse-grainings in which the macroscopic and microscopic state variables always sum to 1.
                                                                                                                \begin{definition}[Semi-Coarsenablity, Coarsenablity, Global Coarsenablity] Let $G, K$ be sets, let $\mathcal W:\Lambda^G\rightarrow\Lambda^G$ be an operator, let $\beta:G\rightarrow K$ be a theme map, and let $U\subseteq\Lambda^G$ such that $\Xi_\beta(U)=\Lambda^K$. We say that $\mathcal W$ is semi-coarsenable under $\beta$ on $U$ if there exists an operator $\mathcal Q:\Lambda^K\rightarrow \Lambda^K$ such that for all
                                                                                                                $p\in U$, $\mathcal Q\circ\Xi_\beta p=\Xi_\beta\circ \mathcal Wp$,
                                                                                                                i.e. the following diagram commutes:
                                                                                                                \[\xymatrix{
                                                                                                                 U \ar[rr]^{\mathcal W} \ar[d]_{\Xi_\beta} && {\Lambda^G}
                                                                                                                 \ar[d]^{\Xi_\beta}\\
                                                                                                                    \Lambda^K\ar[rr]_{\mathcal Q} && \Lambda^K}\]
                                                                                                                Since $\beta$ is surjective, if $\mathcal Q$ exists, it is clearly unique; we call it the quotient. We call $ G, K, W, \text{ and }U$ the domain, co-domain, primary operator and turf respectively. If in addition $\mathcal W(U)\subseteq U$ we say that $\mathcal W$ is coarsenable under $\beta$ on $U$. If in addition $U=\Lambda^G$ we say that $\mathcal W$ is globally coarsenable under $\beta$.
                                                                                                                \end{definition}
                                                                                                                Note that the partition function $\Xi_\beta$ of the coarse-graining is not the same as the theme map $\beta$ of the coarsening.

                                                                                                                Global coarsenablity is a stricter condition than coarsenablity, which in turn is a stricter condition than semi-coarsenablity. It is easily shown that global coarsenablity is equivalent to Vose's notion of compatibility \cite[p. 188]{vose:1999:sgaft} (for a proof see Theorem 17.5 in \citealp{vose:1999:sgaft}).

                                                                                                                If some operator $\mathcal W$ is coarsenable under some theme map $\beta$ on some turf $U$ with some quotient $\mathcal Q$, then for any distribution $\p_K\in\Xi_\beta(U)$, and all distributions $\p_K\ranglel_{\Xi_\beta}$, one can study the \emph{projected} effect of the repeated application of application of $\mathcal W$ to $\p_G$ simply by studying the effect of the repeated application of $\mathcal Q$ to $\p_K$. If the size of $K$ is small then a computational study of the projected effect of the repeated application of $\mathcal W$ to distributions in $U$ becomes feasible.

                                                                                                                \section{Global Coarsenablity of Variation}\label{variationCoarsenability}
                                                                                                                We show that some variation operator $\mathcal V_T$ is globally coarsenable under some theme map if a relationship, that we call \emph{ambivalence}, exists between the transmission function $T$ of the variation operator and the theme map.

                                                                                                                To illustrate the idea of ambivalence consider a theme map  $\beta$ which partitions a genome set $G$ into three subsets. Fig 1 depicts the behavior of a two-parent transmission function that is ambivalent under $\beta$. Given two parents and some child, the probability that the child will belong to some theme class depends \emph{only} on the theme classes of the parents and \emph{not} on the specific parent genomes. Hence the name `ambivalent' --- it captures the sense that when viewed from the coarse-grained level of the theme classes, a transmission function `does not care' about the specific genomes of the parents or the child.

                                                                                                                \begin{figure}[t]\begin{center}
                                                                                                                \includegraphics[height=5.3cm, width=8.5cm]{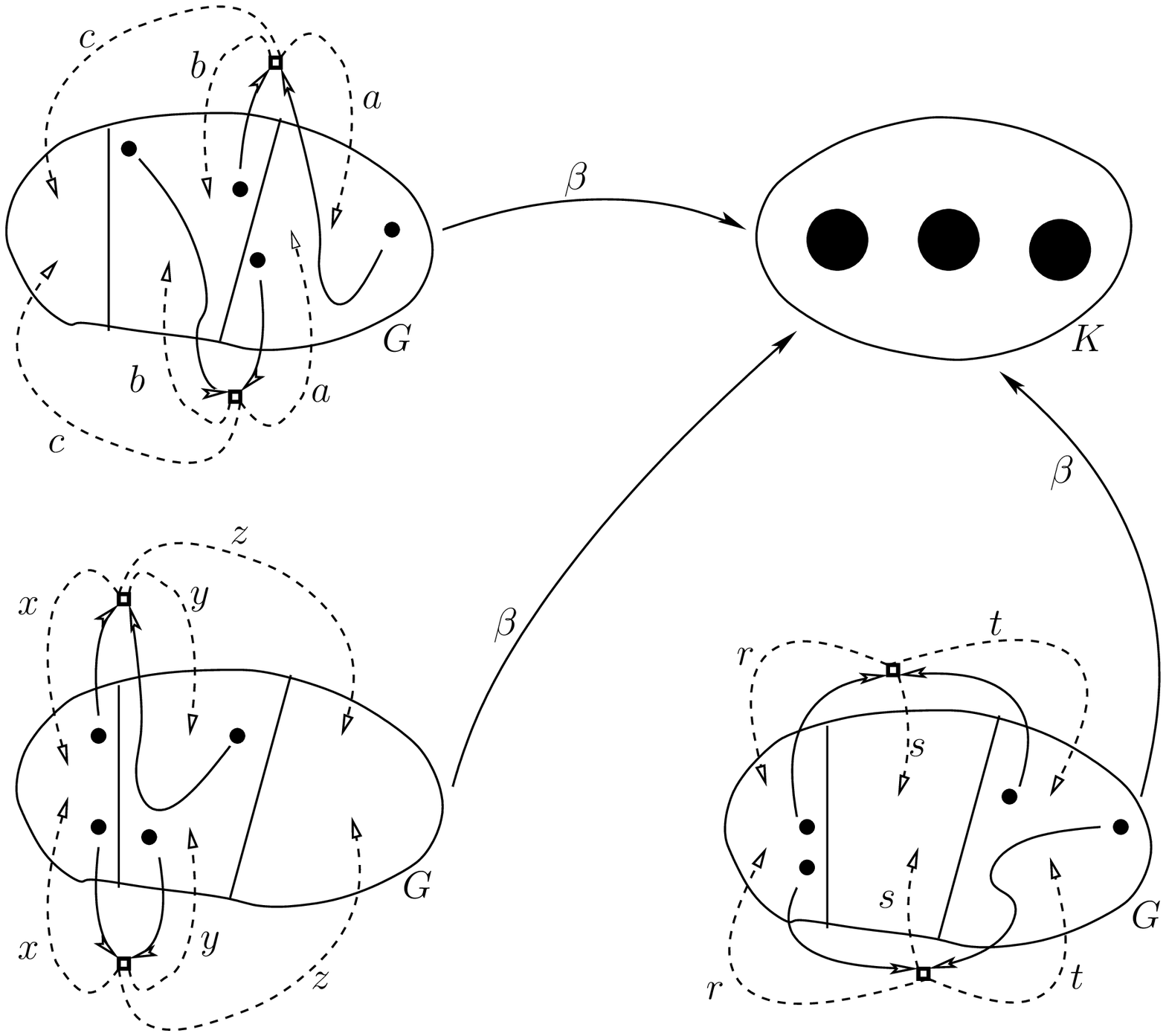}\end{center}
                                                                                                                \caption{Let $\beta:G\rightarrow K$ be a coarse-graining which partitions the genome
                                                                                                                set $G$ into three theme classes. This figure depicts the behavior of a two-parent
                                                                                                                variation operator that is ambivalent under $\beta$. The small dots denote specific
                                                                                                                genomes and the solid unlabeled arrows denote the recombination of these genomes. A dashed
                                                                                                                arrow denotes that a child from a recombination may be produced `somewhere' within the
                                                                                                                theme class that it points to, and the label of a dashed arrow denotes the probability
                                                                                                                with which this might occur. As the diagram shows the probability that the child of a
                                                                                                                variation operation will belong to a particular theme class depends \emph{only} on the
                                                                                                                theme classes of the parents and \emph{not} on their specific genomes}\label{ambivalencfigure}
                                                                                                                \end{figure}

                                                                                                                The definition of ambivalence that follows is equivalent to but more useful than the definition given by \citet{interPopConstaints}
                                                                                                                \begin{definition}\,\,\textsc{(Ambivalence)} Let $G, K$ be countable sets, let  $T\in
                                                                                                                \Lambda^G_{m}$ be a transmission function, and let $\beta:G\rightarrow K$ be a theme map. We say that $T$ is ambivalent under $\beta$ if there exists some transmission function $Y\in
                                                                                                                \Lambda^K_{m}$, such that for all $k,k_1,\ldots, k_m\in K$ and for any $x_1\in\langle
                                                                                                                k_1\rangle_{\!\beta}^{\phantom g},\ldots,x_m\in\langle k_m\rangle_{\!\beta}^{\phantom
                                                                                                                g}$,
                                                                                                                \[\sum_{x\in \langle k
                                                                                                                \rangle_{\!\beta}^{\phantom g}} T(x|x_1,\ldots,x_m)=Y(k|k_1,\ldots, k_m) \] If such a $Y$ exits, it is clearly unique. We denote it by $
                                                                                                                T^{\overrightarrow\beta}$ and call it the theme transmission function.
                                                                                                                \end{definition}

                                                                                                                Suppose $T\in \Lambda^X_{m}$ is ambivalent under some
                                                                                                                $\beta:X\rightarrow K$, we can use the projection operator to express the projection of $T$ under $\beta$ as follows: for all $k, k_1,\ldots, k_m\in K$, and any $x_1\in\langle
                                                                                                                k_1\rangle_{\!\beta}^{\phantom g}, \ldots,x_m\in\langle k_m\rangle_{\!\beta}^{\phantom
                                                                                                                g}$, $T^{\overrightarrow\beta}(k|k_1,\ldots k_m)$ is given by $(\Xi_\beta
                                                                                                                (T(\cdot\,|x_1,\ldots,x_m)))(k)$. The notion of ambivalence is equivalent to a generalization of Toussaint's notion of trivial neutrality \citep[p. 26]{ToussaintThesis}. A one-parent   transmission function is ambivalent under a mapping to the set of phenotypes if and only if it is trivially neutral.

                                                                                                                The following theorem shows that a variation operator is globally coarsenable under some theme map if it is parameterized by a transmission function which is ambivalent under that theme map\ifthenelse{\boolean{showProofs}}{}{\footnote{Due to space
                                                                                                                restrictions proofs have been omitted. They can be found in the full version of this
                                                                                                                paper which is posted on the author's website}}. The method by which we prove this theorem extends the method used by \citet{ToussaintThesis} in his proof of Theorem 1.2.2.

                                                                                                                \begin{theorem}[Global Coarsenablity of Variation] \label{globConcOfVar} Let $G$ and $K$ be countable sets, let
                                                                                                                $T\in \Lambda^G_{m}$ be a transmission function and let  $\beta:G\rightarrow K$ be some
                                                                                                                theme map such that $T$ is ambivalent under $\beta$. Then $\mathcal V_T:\Lambda^G\rightarrow \Lambda^G$ is globally coarsenable under $\beta$ with quotient $\mathcal V_{T^{\overrightarrow\beta}}$, i.e. the following diagram commutes:
                                                                                                                \[\xymatrix{
                                                                                                                 \Lambda^G \ar[rr]^{\mathcal V_T} \ar[d]_{\Xi_\beta} && {\Lambda^G}
                                                                                                                 \ar[d]^{\Xi_\beta}\\
                                                                                                                    \Lambda^K\ar[rr]_{\mathcal V_{T^{\overrightarrow\beta}}} && \Lambda^K}\]
                                                                                                                \end{theorem}
                                                                                                                \ifthenelse{\boolean{showProofs}}{
                                                                                                                \textsc{Proof:  } For any $p\in\Lambda^G$,
                                                                                                                {\allowdisplaybreaks
                                                                                                                \begin{align*}
                                                                                                                &(\Xi_\beta\circ\mathcal V_Tp)(k)\\ &=\sum_{x\in \langle k\rangle_{\!\beta}^{\phantom
                                                                                                                g}}
                                                                                                                \sum_{\substack{(x_1,\ldots,x_m)\\\in\prod\limits_1^{m}X}}T(x|x_1,\ldots,x_m)\prod_{i=1}^mp(x_i)\\
                                                                                                                &=\sum_{\substack{(x_1,\ldots,x_m)\\\in\prod\limits_1^{m}X}}
                                                                                                                \sum_{x\in \langle k\rangle_{\!\beta}^{\phantom g}}T(x|x_1,\ldots,x_m)\prod_{i=1}^mp(x_i)\\
                                                                                                                &=\sum_{\substack{(x_1,\ldots,x_m)\\\in\prod\limits_1^{m}X}}\prod_{i=1}^mp(x_i)
                                                                                                                \sum_{x\in \langle k\rangle_{\!\beta}^{\phantom g}}T(x|x_1,\ldots,x_m)\\
                                                                                                                &=\sum_{\substack{(k_1,\ldots,k_m)\\\in\prod\limits_1^{m}K}}\sum_{\substack{(x_1,\ldots,x_m)\\\in\prod\limits_{j=1}^{m}\langle
                                                                                                                k_j\rangle_{\!\beta}^{\phantom g}}}\prod_{i=1}^mp(x_i) \sum_{x\in \langle k\rangle_{\!\beta}^{\phantom g}}T(x|x_1,\ldots,x_m)\\
                                                                                                                &=\sum_{\substack{(k_1,\ldots,k_m)\\\in\prod\limits_1^{m}K}}\sum_{\substack{(x_1,\ldots,x_m)\\\in\prod\limits_{j=1}^{m}\langle
                                                                                                                k_j\rangle_{\!\beta}^{\phantom g}}}\prod_{i=1}^mp(x_i)
                                                                                                                T^{\overrightarrow \beta}(k|k_1,\ldots,k_m)\\
                                                                                                                &=\sum_{\substack{(k_1,\ldots,k_m)\\\in\prod\limits_1^{m}K}}T^{\overrightarrow
                                                                                                                \beta}(k|k_1,\ldots,k_m)\sum_{\substack{(x_1,\ldots,x_m)\\\in\prod\limits_{j=1}^{m}\langle
                                                                                                                k_j\rangle_{\!\beta}^{\phantom g}}}\prod_{i=1}^mp(x_i)\\
                                                                                                                &=\sum_{\substack{(k_1,\ldots,k_m)\\\in\prod\limits_1^{m}K}}T^{\overrightarrow
                                                                                                                \beta}(k|k_1,\ldots,k_m)\sum_{x_1\in\langle
                                                                                                                k_1\rangle_{\!\beta}^{\phantom g}}\ldots\sum_{x_m\in\langle
                                                                                                                k_m\rangle_{\!\beta}^{\phantom g}}p(x_1)\ldots p(x_m)\\
                                                                                                                &=\sum_{\substack{(k_1,\ldots,k_m)
                                                                                                                \\\in\prod\limits_1^{m}K}}T^{\overrightarrow
                                                                                                                \beta}(k|k_1,\ldots,k_m)\bigg(\sum_{x_1\in\langle
                                                                                                                k_1\rangle}p(x_1)\bigg)\ldots\bigg(\sum_{x_m\in\langle
                                                                                                                k_m\rangle}p(x_m)\bigg)\\
                                                                                                                &=\sum_{\substack{(k_1,\ldots,k_m)\\\in\prod\limits_1^{m}K}}T^{\overrightarrow
                                                                                                                \beta}(k|k_1,\ldots,k_m)\prod_{i=1}^m\bigg((\Xi_\beta p)(k_i)\bigg)\\ &=(\mathcal
                                                                                                                V_{T^{\overrightarrow
                                                                                                                \beta}}\circ\Xi_\beta p)(k)\quad\quad \qed
                                                                                                                \end{align*}
                                                                                                                }}{}

                                                                                                                The so called implicit parallelism theorem \citep{conf/gecco/WrightVR03} is similar to the theorem above. Note however that the former theorem only shows that variation is globally coarsenable if firstly, the genome set consists of ``fixed length strings, where the size of the alphabet can vary from position to position", secondly the partition over the genome set is a schema partition, and thirdly variation is `structural' (see \citealp{conf/gecco/WrightVR03,journals/ec/RoweVW04} for details). The global coarsenablity of variation theorem has none of these specific requirements. Instead it is premised on the existence of an abstract relationship---ambivalence---between the variation operation and a theme map. The abstract nature of this relationship makes this theorem applicable to evolutionary algorithms other than GAs. In addition this theorem illuminates the essential relationship between `structural' variation and schemata which was used (implicitly) in the proof of the implicit parallelism theorem of \citet{conf/gecco/WrightVR03}.

                                                                                                                In sections \ref{AlgebraOfAmbivalence} and \ref{simVarConsts} we show that a transmission function that models any  combination of variation operations that are commonly used in GAs---i.e. any combination of mask based crossover and canonical mutation, in any order---is  ambivalent under any schema map (defined in definition \ref{defSchemaMap}). Therefore any combination of common variation operators with canonical mutation is globally coarsenable under \emph{any} schema map. This is equivalent to the result of the implicit parallelism theorem.

                                                                                                                \section{Limitwise Semi-Coarsenablity of Selection}\label{selectionCoarsenability}
                                                                                                                For some fitness function $f:G\rightarrow \mathbb R^+$ and some theme map  $\beta:G\rightarrow K$ let us say that $f$ is \emph{thematically invariant} under $\beta$ if, for any thematically $k\in K$, the genomes that belong to $\langle k \rangle_\beta$ all have the same fitness.
                                                                                                                Paraphrasing the comments of \citet{conf/gecco/WrightVR03} using the terminology developed in this paper, Wright et. al. argue that if the selection operator  is globally coarsenable under some schema map $\beta:G\rightarrow K$ then the fitness function that parameterizes the selection operator is `schematically' invariant under $\beta$. It is relatively simple to use contradiction to prove a generalization of this statement for arbitrary theme maps.

                                                                                                                Schematic invariance is a very strict condition for a fitness function. An IPSGA whose fitness function meets this condition is unlikely to yield any substantive information about the dynamics of real world GAs.

                                                                                                                As stated above, the selection operator is not \emph{globally} coarsenable unless the fitness function satisfies thematic invariance, however if the set of distributions that selection operates over (i.e. the turf) is appropriately constrained, then, as we show in this section, the selection operator is \emph{semi-}coarsenable over the turf even when the fitness function only satisfies a much \emph{weaker} condition called thematic \emph{mean} invariance.\\

                                                                                                                For any theme map $\beta:G\rightarrow K$, any theme $k$, and any distribution $p\in\Lambda^G$, the theme conditional operator, defined below, returns a conditional distribution in  $\Lambda^G$ that is obtained by normalizing the probability mass of the elements in $\langle k\rangle_\beta$ by $(\Xi_\beta p)(k)$

                                                                                                                \begin{definition}[Theme Conditional Operator] Let $G$ be some countable set, let $K$ be some set, and let $\beta:G\rightarrow K$ be some function. We define the theme conditional operator $\mathcal C_\beta:\Lambda^G\times K\rightarrow \Lambda^G\cup 0^G$ as follow: For any $p\in\Lambda^G$, and any $k\in K$, $\mathcal C_\beta(p,k)\in\Lambda^G\cup 0^G$ such that for any $x\in \langle k \rangle_\beta$,
                                                                                                                \[(\mathcal C_\beta(p,k))(x)=\left\{\begin{array}{cl}0&\text{if } \beta(x)\not=k \text{ or }(\Xi_\beta p)(k)=0\\\frac{p(x)}{(\Xi_\beta p)(k)}& \text{otherwise}\end{array}\right.\]
                                                                                                                \end{definition}

                                                                                                                A useful property of the theme conditional operator is that it can be composed with the expected fitness operator to give an operator that returns the average fitness of the genomes in some theme class. To be precise, given some finite genome set $G$, some theme map $\beta:G\rightarrow K$, some fitness function $f:G\rightarrow \mathbb R^+$, some distribution $p\in \Lambda^G$, and some theme $k\in K$, $\mathcal E_f\circ\mathcal C_\beta(p,k)$ is the average fitness of the genomes in $\langle k \rangle_\beta$. This property proves useful in the following definition.

                                                                                                                \begin{definition}[Bounded Thematic Mean Divergence, Thematic Mean Invariance] Let $G$ be some finite set, let $K$ be some set, let $\beta:G\rightarrow K$ be a theme map, let $f:G\rightarrow\mathbb R^+$ and $f^*:K\rightarrow \mathbb R^+$ be functions, let $U\subseteq\Lambda^G$, and let $\delta\in \mathbb R_0^+$. We say that the thematic mean divergence of $f$ with respect to $f^*$ on $U$ under $\beta$ is bounded by $\delta$ if, for any $p\in U$ and for any $k\in K$
                                                                                                                \[|\mathcal E_f\circ\mathcal C_\beta(p, k)-f^*(k)|\leq\delta\]
                                                                                                                \end{definition}

                                                                                                                The next definition gives us a means to measure a `distance' between real valued functions over finite sets.
                                                                                                                \begin{definition}[Manhattan Distance Between Real Valued Functions] Let $X$ be a finite set then for any functions $f,h$ of type $X\rightarrow \mathbb R$ we define the manhattan distance between $f$ and $h$, denoted by $d(f,h)$, as follows:
                                                                                                                \[d(f,h)=\sum_{x\in X}|f(x)-h(x)|\]
                                                                                                                \end{definition}
                                                                                                                \noindent It is easily checked that $d$ is a metric.

                                                                                                                Let $f:G\rightarrow \mathbb R^+$, $\beta:G\rightarrow K$ and $f^*:K\rightarrow \mathbb R^+$ be functions with finite domains, and let $U\in \Lambda^G$. The following theorem shows that if the thematic mean divergence of $f$ with respect to $f^*$ on $U$ under $\beta$ is bounded by some $\delta$, then in the limit as $\delta\rightarrow 0$, $\mathcal S_f$ is semi-coarsenable under $\beta$ on  $U$ .

                                                                                                                \begin{theorem}[Limitwise Semi-Coarsenablity of Selection]\label{SelectionUnderCoarsegraining}
                                                                                                                Let $G$ and $K$ be finite sets, let $\beta:G\rightarrow K$ be a theme map,  Let $U\subseteq\Lambda^G$ such that $\Xi_\beta(U)=\Lambda^K$, let $f:G\rightarrow \mathbb R^+$, $f^*:K\rightarrow \mathbb R^+$ be some functions such that the thematic mean divergence of $f$ with respect to $f^*$ on $U$ under $\beta$ is bounded by some $\delta>0$, then for any $p\in U$, \[\lim\limits_{\delta\rightarrow0} d( \Xi_\beta \circ \mathcal S_fp, \mathcal S_{f^*}\circ \Xi_\beta p)=0\]

                                                                                                                \end{theorem}
                                                                                                                \noindent We depict the result of this theorem as follows:
                                                                                                                \[\xymatrix{
                                                                                                                 U \ar[rr]^{\mathcal S_f} \ar[d]_{\Xi_\beta} \ar @{}[rrd] | {\lim\limits_{\delta\rightarrow 0}}&& {\Lambda^G}
                                                                                                                 \ar[d]^{\Xi_\beta} \\
                                                                                                                    \Lambda^K\ar[rr]_{\mathcal S_{f^*}} && \Lambda^K}\]

                                                                                                                \ifthenelse{\boolean{showProofs}}{
                                                                                                                \noindent \textsc{Proof:  } Let  $\epsilon>0$. We prove that there exists a $\delta'>0$ such that \[\delta<\delta'\Rightarrow d(\Xi_\beta \circ \mathcal S_fp, \mathcal S_{f^*}\circ \Xi_\beta p)<\epsilon\]
                                                                                                                For any $k \in K$,
                                                                                                                {\allowdisplaybreaks
                                                                                                                \begin{align*}
                                                                                                                &(\Xi_\beta\circ\mathcal S_{f}p)(k)\\&=\sum_{g\in\langle k
                                                                                                                \rangle_{\!\beta}^{\phantom g}}(\mathcal S_{f}p)(g)\\
                                                                                                                &=\sum_{g\in\langle k\rangle_{\!\beta}^{\phantom g}}\frac{f(g).p(g)}{\sum_{g'\in G}f(g').p(g')}
                                                                                                                \\&=\frac{\sum\limits_{g\in\langle
                                                                                                                k\rangle_{\!\beta}^{\phantom g}}f(g).(\Xi_\beta p)(k).
                                                                                                                (\mathcal C_\beta(p,k))(g)}{\sum\limits_{k'\in K}\sum\limits_{g'\in
                                                                                                                \langle k' \rangle_{\!\beta}^{\phantom g}} f(g').(\Xi_\beta p)(k')(\mathcal C_\beta(p,k'))(g')} \\
                                                                                                                &=\frac{(\Xi_\beta p)(k)\sum\limits_{g\in\langle
                                                                                                                k\rangle_{\!\beta}^{\phantom g}}f(g).(\mathcal C_\beta(p,k))(g)}{\sum\limits_{k'\in K}(\Xi_\beta p)(k')\sum\limits_{g'\in \langle
                                                                                                                k'\rangle_{\!\beta}^{\phantom g}}
                                                                                                                f(g').(\mathcal C_\beta(p,k'))(g')}\\
                                                                                                                &=\frac{(\Xi_\beta p)(k).\mathcal E_f\circ\mathcal C_\beta(p,k)}{\sum\limits_{k'\in K}(\Xi_\beta p^{\phantom
                                                                                                                t}_G)(k').\mathcal E_f\circ\mathcal C_\beta(p,k')}\\
                                                                                                                &=(\mathcal S_{\mathcal E_f\circ\mathcal C_\beta(p,\cdot)}\circ\Xi_\beta p)(k)
                                                                                                                \end{align*}
                                                                                                                So we have that \[ d(\Xi_\beta \circ\mathcal S_fp, \mathcal S_{f^*}\circ\Xi_\beta p) = d(\mathcal S_{\mathcal E_f\circ\mathcal C_\beta(p,\cdot)}\circ\Xi_\beta p, \mathcal S_{f^*}\circ\Xi_\beta p)\]

                                                                                                                \noindent By Lemma \ref{lemmaD} (in the appendix)  there exists a $\delta_1>0$ such that,
                                                                                                                \[d(\mathcal E_f\circ\mathcal C_\beta(p,.),f^*)<\delta_1\Rightarrow d(\mathcal S_{\mathcal E_f\circ\mathcal C_\beta(p,\cdot)}(\Xi_\beta p), \mathcal S_{f^*}(\Xi_\beta p))<\epsilon\]

                                                                                                                \noindent Let $\delta'=\frac{\delta_1}{|K|}$. Now, if $\delta<\delta'$, then $d(\mathcal E_f\circ\mathcal C_\beta(p,.),f^*)<\delta_1$, so $ d(\Xi_\beta \circ \mathcal S_fp, \mathcal S_{f^*}\circ \Xi_\beta p)<\epsilon \quad\quad \qed$
                                                                                                                }}{}
                                                                                                                \\
                                                                                                                The proof of the following theorem should be obvious from the proof of the preceding one. We therefore omit it.
                                                                                                                \begin{theorem} \label{easySelectionCoarsenability} Let $G$ and $K$ be finite sets, let $\beta:G\rightarrow K$ be a theme map, let $f:G\rightarrow \mathbb R^+, f^*:K\rightarrow\mathbb R^+$ be functions such that for any $k\in K$, and for any $g\in\langle k \rangle_\beta$, $f(g)=f^*(k)$.  $S_f$ is globally coarsenable under $\beta$ with quotient $\mathcal S_{f^*}$, i.e. the following diagram commutes:
                                                                                                                \[\xymatrix{
                                                                                                                 \Lambda^G \ar[rr]^{\mathcal S_f} \ar[d]_{\Xi_\beta} && {\Lambda^G}
                                                                                                                 \ar[d]^{\Xi_\beta} \\
                                                                                                                    \Lambda^K\ar[rr]_{\mathcal S_{f^*}} && \Lambda^K}\]\end{theorem}

                                                                                                                \section{Limitwise Coarsenablity of Evolution}\label{evolutionCoarsenability}

                                                                                                                The two definitions below formalize the idea of an infinite population model of an EA, and its dynamics\footnote{The definition of an Evolution Machine given here is different from that given by \citet{Burjorjee05-ThRepWkshp,conf/iicai/BurjorjeeP05}. The fitness function in this definition maps genomes directly to fitness values. It therefore subsumes the genome-to-phenotype and the phenotype-to-fitness functions of the previous definition. In previous work these two functions were always composed together; their subsumption within a single function increases clarity.}.

                                                                                                                \begin{definition}[Evolution Machine] An evolution machine (EM) is a tuple $(G, T, f)$ where $G$ is some set called the domain, $f:G\rightarrow \mathbb R^+$ is a function called the fitness function and $T\in \Lambda^G_m$ is called the transmission function.
                                                                                                                \end{definition}

                                                                                                                \begin{definition}[Evolution Epoch Operator] Let $E=(G,T,f)$ be an evolution machine. We define the evolution epoch operator $\mathcal G_E:\Lambda^G\rightarrow \Lambda^G$ as follows:
                                                                                                                \[\mathcal G_E=\mathcal V_T\circ\mathcal S_f\]\end{definition}

                                                                                                                \noindent The following theorem follows easily from theorems \ref{variationCoarsenability} and \ref{easySelectionCoarsenability}.

                                                                                                                \begin{theorem} Let $E=(G,T,f), E^*=(K, T^*, f^*)$ be evolution machines such that $G$ and $K$ are finite.  Let $\beta:G\rightarrow K$ be a theme map. For any $k\in K$, and for any $g\in\langle k \rangle_\beta$, let $f(g)=f^*(k)$.  let $T^*=T^{\overrightarrow{\beta}}$. Then $\mathcal G_E$ is globally coarsenable under $\beta$, i.e. the following diagram commutes:

                                                                                                                \[\xymatrix{
                                                                                                                 \Lambda^G \ar[rr]^{\mathcal G_E} \ar[d]_{\Xi_\beta} && {\Lambda^G}
                                                                                                                 \ar[d]^{\Xi_\beta} \\
                                                                                                                    \Lambda^K\ar[rr]_{\mathcal G_{E^*}} && \Lambda^K}\]

                                                                                                                \end{theorem}
                                                                                                               The condition that $f$ be constant over each theme class is too severe for this theorem to be of any use in practice. We seek a coarsenability result where $f$ is less severely constrained.

                                                                                                                \begin{definition}[Non-Departure] Let $E=(G,T,f)$ be an evolution machine, let $U\subseteq\Lambda^G$, and let $p\in U$. For any $\tau\in\mathbb Z^+$, we say that $E$ on $p$ is \emph{non-departing} from  $U$ for $\tau$ generations if for all $t\in\mathbb Z^+$ such that $t\leq\tau$,
                                                                                                                \[\mathcal G_E^t(p)\in U\]
                                                                                                                \end{definition}

                                                                                                                \begin{theorem}[Limitwise Coarsenablity of Evolution] \label{limCorsEvol}
                                                                                                                Let $E=(G,T,f)$, be an evolution machine such that $G$ is finite, let $\beta:G\rightarrow K$ be some theme map, let $f^*:K\rightarrow \mathbb R^+$ be some function, let $\delta\in\mathbb R^+_0$, let $U\subseteq\Lambda^G$ such that $\Xi_\beta(U)=\Lambda^K$, let $p\in U$, and let $\tau\in\mathbb Z^+$. Suppose that the following statements are true:
                                                                                                                \begin{enumerate}
                                                                                                                \item The thematic mean divergence of $f$ with respect to $f^*$ on $U$ under $\beta$ is bounded by $\delta$
                                                                                                                \item $T$ is ambivalent under $\beta$
                                                                                                                \item $E$ on $p$ is non-departing from $U$ for $\tau$ generations
                                                                                                                \end{enumerate}
                                                                                                                Then, letting $E^*=(K, T^{\overrightarrow{\beta}}, f^*)$ be an evolution machine, for any $t\in \mathbb Z^+_0$ such that $t\leq\tau$, we have that
                                                                                                                \[\lim\limits_{\delta\rightarrow0} d(\Xi_\beta\circ \mathcal G^t_Ep\,\,,\,\, \mathcal G^t_{E^*}\circ\Xi_\beta p)=0\]

                                                                                                                \end{theorem}

                                                                                                                \noindent where the equation above is depicted as follows:
                                                                                                                \[\xymatrix{
                                                                                                                 U \ar[rr]^{\mathcal G^t_E} \ar[d]_{\Xi_\beta} \ar @{}[rrd] | {\lim\limits_{\delta\rightarrow 0}}&& U
                                                                                                                 \ar[d]^{\Xi_\beta}\\
                                                                                                                    \Lambda^K\ar[rr]_{\mathcal G^t_{E^*}} && \Lambda^K}\]
                                                                                                                \ifthenelse{\boolean{showProofs}}{
                                                                                                                {\sc Proof:  }
                                                                                                                For any $t\leq\tau$ we prove that for any $\epsilon>0$ there exists a $\delta'>0$ such that
                                                                                                                \[\delta<\delta' \Rightarrow d(\Xi_\beta\circ \mathcal G^t_Ep\,\,,\,\, \mathcal G^t_{E^*}\circ\Xi_\beta p)<\epsilon\]
                                                                                                                The proof is by induction on $t$. The base case, when $t=0$, is trivial. For some $n=\mathbb Z^+_0$, such that $n<\tau$, let us assume the hypothesis for $t=n$. We now show that it is true for $t=n+1$. Note that,


                                                                                                                \begin{align*}
                                                                                                                &d(\Xi_\beta\circ\mathcal G^{n+1}_Ep\, , \, \mathcal G^{n+1}_{E^*}\circ\Xi_\beta\p)\\
                                                                                                                &=d(\Xi_\beta\circ\mathcal V_T\circ\mathcal S_f\circ\mathcal G^{n}_Ep\, , \, \mathcal V_{T^{\overrightarrow\beta}}\circ\mathcal S_{f^*}\circ\mathcal G^{n}_{E^*}\circ\Xi_\beta\p)\\
                                                                                                                &=d(\mathcal V_{T^{\overrightarrow\beta}}\circ\Xi_\beta\circ\mathcal S_f\circ\mathcal G^{n}_Ep\, , \, \mathcal V_{T^{\overrightarrow\beta}}\circ\mathcal S_{f^*}\circ\mathcal G^{n}_{E^*}\circ\Xi_\beta p)&\text{(by theorem \ref{globConcOfVar})}
                                                                                                                \end{align*}

                                                                                                                \noindent Hence, for any $\epsilon>0$, by Lemma \ref{lemmaB} there exists $\delta_1$ such that
                                                                                                                \begin{align*}d(\Xi_\beta\circ\mathcal S_f\circ\mathcal G^{n}_Ep\, , \, \mathcal \mathcal S_{f^*}\circ\mathcal G^{n}_{E^*}\circ\Xi_\beta p)<\delta_1\Rightarrow  d(\Xi_\beta\circ\mathcal G^{n+1}_Ep\, , \, \mathcal G^{n+1}_{E^*}\circ\Xi_\beta\p)<\epsilon\end{align*}
                                                                                                                \noindent As $d$ is a metric it satisfies the triangle inequality. Therefore we have that
                                                                                                                \begin{multline*}
                                                                                                                d(\Xi_\beta\circ\mathcal S_f\circ\mathcal G^{n}_Ep\, , \, \mathcal S_{f^*}\circ\mathcal G^{n}_{E^*}\circ\Xi_\beta p)\leq\\
                                                                                                                d(\Xi_\beta\circ\mathcal S_{f}\circ \mathcal G^{n}_Ep\, , \, \mathcal S_{f^*}\circ\Xi_\beta\circ\mathcal G^{n}_{E} p)+\\d(\mathcal S_{f^*}\circ\Xi_\beta\circ\mathcal G^{n}_{E} p\, , \, \mathcal S_{f^*}\circ\mathcal G^{n}_{E^*}\circ\Xi_\beta p)
                                                                                                                \end{multline*}
                                                                                                                By the definition of departure,  $\mathcal G_E^np\,\,\in U$. So, by  theorem \ref{SelectionUnderCoarsegraining} there exists a $\delta_2$ such that
                                                                                                                \[\delta<\delta_2\Rightarrow d(\Xi_\beta\circ\mathcal S_{f}\circ \mathcal G^{n}_Ep\, , \, \mathcal S_{f^*}\circ\Xi_\beta\circ\mathcal G^{n}_{E} p)<\frac{\delta_1}{2}\]

                                                                                                                \noindent By lemma \ref{lemmaC} there exists a $\delta_3$ such that
                                                                                                                \[d(\Xi_\beta\circ\mathcal G^{n}_{E} p\, , \, \mathcal G^{n}_{E^*}\circ\Xi_\beta p)<\delta_3\Rightarrow d(\mathcal S_{f^*}\circ\Xi_\beta\circ\mathcal G^{n}_{E} p\, , \, \mathcal S_{f^*}\circ\mathcal G^{n}_{E^*}\circ\Xi_\beta p)<\frac{\delta_1}{2}\]

                                                                                                                \noindent By our inductive assumption, there exists a $\delta_4$ such that \[\delta<\delta_4 \Rightarrow  d(\Xi_\beta\circ\mathcal G^{n}_{E} p\, , \, \mathcal G^{n}_{E^*}\circ\Xi_\beta p)<\delta_3\]

                                                                                                                \noindent Therefore, letting $\delta'=\text{min}(\delta_2, \delta_4)$ we get that
                                                                                                                \[\delta<\delta'\Rightarrow d(\Xi_\beta\circ \mathcal G^{n+1}_Ep, \mathcal G^{n+1}_{E^*}\circ\Xi_\beta p)<\epsilon \quad\quad\qed\]}{}

                                                                                                                The limitwise coarsenability of evolution theorem is very general. As we have not committed ourselves to any particular genomic data-structure the coarse-graining result we have obtained is applicable to any IPEA provided that it satisfies three abstract conditions: bounded thematic mean divergence, ambivalence, and non-departure. The fidelity of the coarse-graining depends on the the minimal bound on the thematic mean divergence. Maximum fidelity is achieved in the limit as this minimal bound tends to zero.

                                                                                                                Note that apart from the way it is parameterized, the quotient operator is the same as the primary operator. We therefore say that the coarse-graining is \emph{operationally invariant}. This feature is very significant. It means that the frequency dynamics of the themes of an IPEA which satisfies the three abstract conditions mentioned above, can be accurately predicted by the frequency dynamics of the genomes of another IPEA, for some number of generations, provided that the minimal bound of the thematic mean divergence of the former IPEA is sufficiently low.

                                                                                                                \section{The Algebra of Ambivalence} \label{AlgebraOfAmbivalence}

                                                                                                                Given several transmission functions $T_1$, \ldots $T_n$ that are all ambivalent under
                                                                                                                some theme map $\beta$ we show two ways of combining $T_1$, \ldots $T_n$ to create
                                                                                                                a new transmission function that is also ambivalent under $\beta$. Also, given a single
                                                                                                                transmission function $T$ and several theme maps $\beta_1, \ldots, \beta_n$ such
                                                                                                                that $T$ is ambivalent under each of these theme maps, we show how, if a certain
                                                                                                                condition is met, the theme maps can be combined to create a new theme map
                                                                                                                such that $T$ will be ambivalent under this new theme map.

                                                                                                                \subsection{Combining Transmission functions}
                                                                                                                Given several ambivalent transmission functions over some set, the following two lemmas
                                                                                                                give us two ways to create new ambivalent transmission functions --- firstly by taking a
                                                                                                                weighted sum, and secondly by composition.
                                                                                                                \begin{lemma},\,\,\ \textsc{(The Weighted sum of Ambivalent Transmission Functions is Ambivalent)}
                                                                                                                \label{WeightedSumAmbivalent}For any set $X$, any function $\beta:X\rightarrow K$, and
                                                                                                                any $n\in\mathbb N^+$, let $T_1,\ldots, T_n\in \Lambda^X_m$ be ambivalent transmission
                                                                                                                functions. Let $p\in \Lambda^{\{1,\ldots,n\}}$ and let $T\in \Lambda^X_m$ be defined as
                                                                                                                follows:
                                                                                                                \[T(x|x_1\ldots,x_m)=\sum_{i=1}^np(i)\,T_i(x|x_1,\ldots,x_m)\]
                                                                                                                Then $T$ is ambivalent under $\beta$ with $\beta$-projection $T^{\overrightarrow\beta}$
                                                                                                                given as follows: \[T^{\overrightarrow\beta}(k|k_1,\ldots,
                                                                                                                k_m)=\sum_{i=1}^np(i)\,T_i^{\overrightarrow \beta}(k|k_1,\ldots, k_m)\]
                                                                                                                \end{lemma}\ifthenelse{\boolean{showProofs}}{{\sc Proof:  }
                                                                                                                For any $x_1\in\langle
                                                                                                                k_1\rangle_{\!\beta}^{\phantom g},\ldots,x_m\in\langle k_m\rangle_{\!\beta}^{\phantom
                                                                                                                g}$\begin{align*}T^{\overrightarrow
                                                                                                                \beta}(k|k_1,\ldots,k_m)&=\sum_{x\in\langle k\rangle_{\!\beta}^{\phantom g}}T(x|x_1,\ldots,x_m)\\
                                                                                                                &=\sum_{x\in\langle k\rangle_{\!\beta}^{\phantom g}}\sum_{i=1}^np(i)T_i(x|x_1,\ldots,x_m)\\
                                                                                                                &=\sum_{i=1}^np(i)\sum_{x\in\langle k\rangle_{\!\beta}^{\phantom g}}T_i(x|x_1,\ldots,x_m)\\
                                                                                                                &=\sum_{i=1}^np(i)T^{\overrightarrow\beta}(k|k_1,\ldots,k_m) \quad\quad\qed
                                                                                                                \end{align*}}{}

                                                                                                                \begin{lemma}\,\,\,\textsc{(The Composition of Ambivalent Transmission Functions is Ambivalent)} \label{CompositionAmbivalent}For any $T_1\in{\Lambda^X_m}, T_2\in{\Lambda^X_n}$, if $T_1$ and $T_2$ are both
                                                                                                                ambivalent under some theme map $\beta:X\rightarrow K$. Then $T_1\circ T_2$ is
                                                                                                                ambivalent under $\beta$ with $\beta$-projection given as follows:
                                                                                                                \begin{align*}(T_1\circ
                                                                                                                T_2)^{\overrightarrow\beta}(k|j_1,\ldots,j_n)=\sum_{\substack{(k_1,\ldots,k_m)\\\in\Pi_1^m
                                                                                                                K}} T_1^{\overrightarrow \beta}(k|k_1,\ldots,k_m)\prod_{i=1}^mT_2^{\overrightarrow
                                                                                                                \beta}(k_i|j_1,\ldots,j_n)\end{align*}
                                                                                                                \end{lemma}\ifthenelse{\boolean{showProofs}}{{\sc Proof:  }
                                                                                                                For any $y_1\in\langle j_1\rangle_{\!\beta}^{\phantom g} ,\ldots, y_n\in\langle j_n\rangle_{\!\beta}^{\phantom g}$,
                                                                                                                {\allowdisplaybreaks
                                                                                                                \begin{align*}
                                                                                                                &(T_1\circ T_2)^{\overrightarrow\beta}(k|j_1,\ldots,j_n)\\
                                                                                                                &=\sum_{x\in \langle k\rangle_{\!\beta}^{\phantom
                                                                                                                g}}\,\,\,\sum_{\substack{(x_1,\ldots,x_m)\in\Pi_1^m
                                                                                                                X}}T_1(x|x_1,\ldots,x_m)\prod_{i=1}^mT_2(x_i|y_1,\ldots,y_n)\\
                                                                                                                &=\sum_{x\in \langle k\rangle_\beta}\,\,\,\sum_{\substack{(k_1,\ldots,k_m)\in\Pi_1^m
                                                                                                                K}}\,\,\,\sum_{\substack{(x_1,\ldots,x_m)\in\Pi_{\ell=1}^m \langle k_\ell
                                                                                                                \rangle_{\!\beta}^{\phantom
                                                                                                                g}}}T_1(x|x_1,\ldots,x_m)\prod_{i=1}^mT_2(x_i|y_1,\ldots,y_n)\\
                                                                                                                &=\sum_{x\in \langle k\rangle_\beta}\,\,\,\sum_{\substack{(k_1,\ldots,k_m)\in\Pi_1^m
                                                                                                                K}}\,\,\,\sum_{\substack{(x_1,\ldots,x_m)\in\Pi_{\ell=1}^m \langle k_\ell
                                                                                                                \rangle_{\!\beta}^{\phantom
                                                                                                                g}}}\prod_{i=1}^mT_2(x_i|y_1,\ldots,y_n)T_1(x|x_1,\ldots,x_m)\\
                                                                                                                &=\sum_{x\in \langle k\rangle_{\!\beta}^{\phantom
                                                                                                                g}}\sum_{\substack{(k_1,\ldots,k_m)\in\Pi_1^mK}} \,\,\,\sum_{x_1 \in\langle k_1
                                                                                                                \rangle_{\!\beta}^{\phantom g}}T_2(x_1|y_1,\ldots,y_n)\sum_{x_2 \in\langle k_2
                                                                                                                \rangle_{\!\beta}^{\phantom g}}T_2(x_2|y_1,\ldots,y_n)\\
                                                                                                                &\indent\indent \sum_{x_3 \in\langle k_3
                                                                                                                \rangle_{\!\beta}^{\phantom g}}T_2(x_3|y_1,\ldots,y_n)\ldots\sum_{x_m \in \langle k_m
                                                                                                                \rangle_{\!\beta}^{\phantom g}}T_2(x_m|y_1,\ldots,y_n)T_1(x|x_1,\ldots,x_m)\\
                                                                                                                &=\sum_{\substack{(k_1,\ldots,k_m)\\\in\Pi_1^mK}} \,\,\,\sum_{x_1 \in\langle k_1
                                                                                                                \rangle_{\!\beta}^{\phantom g}}T_2(x_1|y_1,\ldots,y_n)\sum_{x_2 \in\langle k_2
                                                                                                                \rangle_{\!\beta}^{\phantom g}}T_2(x_2|y_1,\ldots,y_n)\\
                                                                                                                &\indent\indent\sum_{x_3 \in\langle k_3
                                                                                                                \rangle_{\!\beta}^{\phantom g}}T_2(x_3|y_1,\ldots,y_n)\ldots\sum_{x_m \in \langle k_m
                                                                                                                \rangle_{\!\beta}^{\phantom g}}T_2(x_m|y_1,\ldots,y_n)\sum_{x\in \langle
                                                                                                                k\rangle_{\!\beta}^{\phantom g}}T_1(x|x_1,\ldots,x_m)\\
                                                                                                                &=\sum_{\substack{(k_1,\ldots,k_m)\\\in\Pi_1^mK}} \sum_{x_1 \in\langle k_1
                                                                                                                \rangle_{\!\beta}^{\phantom g}}T_2(x_1|y_1,\ldots,y_n)\sum_{x_2 \in\langle k_2
                                                                                                                \rangle_{\!\beta}^{\phantom g}}T_2(x_2|y_1,\ldots,y_n)\\
                                                                                                                &\indent\indent \sum_{x_3 \in\langle k_3
                                                                                                                \rangle_{\!\beta}^{\phantom g}}T_2(x_3|y_1,\ldots,y_n)\ldots\sum_{x_m \in \langle k_m
                                                                                                                \rangle_{\!\beta}^{\phantom
                                                                                                                g}}T_2(x_m|y_1,\ldots,y_n)T_1^{\overrightarrow\beta}(k|k_1,\ldots,k_m)\\
                                                                                                                &=\sum_{\substack{(k_1,\ldots,k_m)\in\Pi_1^mK}}
                                                                                                                T_1^{\overrightarrow\beta}(k|k_1,\ldots,k_m)\sum_{x_1 \in\langle k_1
                                                                                                                \rangle_{\!\beta}^{\phantom g}}T_2(x_1|y_1,\ldots,y_n)\ldots\sum_{x_m \in\langle
                                                                                                                k_m \rangle_{\!\beta}^{\phantom g}}T_2(x_m|y_1,\ldots,y_n)\\
                                                                                                                &=\sum_{\substack{(k_1,\ldots,k_m)\\\in\Pi_1^mK}}
                                                                                                                T_1^{\overrightarrow\beta}(k|k_1,\ldots,k_m)\bigg(\sum_{x_1 \in\langle k_1
                                                                                                                \rangle_{\!\beta}^{\phantom g}}T_2(x_1|y_1,\ldots,y_n)\bigg)\ldots\bigg(\sum_{x_m
                                                                                                                \langle k_m \rangle_{\!\beta}^{\phantom g}}T_2(x_m|y_1,\ldots,y_n)\bigg)\\
                                                                                                                &=\sum_{\substack{(k_1,\ldots,k_m)\\\in\Pi_1^mK}}
                                                                                                                T_1^{\overrightarrow\beta}(k|k_1,\ldots,k_m)\prod_{i=1}^mT_2^{\overrightarrow\beta}(k_i|j_1,\ldots,j_n)\quad\quad\qed
                                                                                                                \end{align*}}}{}

                                                                                                                \subsection{Combining Theme maps} Given several theme maps with the same domain, the following definition allows
                                                                                                                us to create a new theme map which induces a finer partition over the domain.
                                                                                                                \begin{definition}\,\,\,\textsc{(Cartesian Product of Theme maps)}
                                                                                                                For any $n\in \mathbb N^+$ and any functions $\beta_1:X\rightarrow K_1$,\ldots,
                                                                                                                $\beta_n:X\rightarrow K_n$, which share the same domain we define the cartesian product
                                                                                                                of $\beta_1,\ldots,\beta_n$ to be the function
                                                                                                                $\beta_1\times\ldots\times\beta_n:X\rightarrow \prod_{i=1}^nK_i$ as follows:
                                                                                                                \[(\beta_1\times\ldots\times\beta_n)(x)=(\beta_1(x),\ldots,\beta_n(x))\]
                                                                                                                \end{definition}
                                                                                                                For notational convenience we will denote some cartesian product
                                                                                                                $\beta_1\times\ldots\times\beta_n$ as $\prod_{i=1}^n\beta_i$.

                                                                                                                Given two theme maps with the same domain $\beta_1:X\rightarrow K_1$ and
                                                                                                                $\beta_2:X\rightarrow K_2$ and some transmission function $T\in \Lambda^X_m$, we say that
                                                                                                                $\beta_1$ and $\beta_2$ are independent with respect to $T$ if for any choice of $m$
                                                                                                                parents $x_1,x_2,\ldots,x_m$, the mutual information between
                                                                                                                $\Xi_{\beta_1}(T(\cdot|x_1,\ldots,x_m))$ and $\Xi_{\beta_2}(T(\cdot|x_1,\ldots,x_m))$ is
                                                                                                                zero. In other words knowing something about the distribution
                                                                                                                $\Xi_{\beta_1}(T(\cdot|x_1,\ldots,x_m))$ gives no information about the distribution
                                                                                                                $\Xi_{\beta_2}(T(\cdot|x_1,\ldots,x_m))$. Formally, for all $k_1\in K_1$, $k_2\in K_2$
                                                                                                                \begin{align*}\Xi_{\beta_1\times\beta_2}(T(\cdot|x_1,\ldots,x_m))(k_1,k_2)=
                                                                                                                \Xi_{\beta_1}(T(\cdot|x_1,\ldots,x_m))(k_1)\Xi_{\beta_2}(T(\cdot|x_1,\ldots,x_m))(k_2)\end{align*}

                                                                                                                The following definition extends this idea of independence to multiple theme maps.
                                                                                                                Unfortunately it defines independence in a form that is more suited for use in proofs
                                                                                                                than for understandability. The proposition that follows shows that this form is
                                                                                                                equivalent to the more intuitive form that we used above.

                                                                                                                \begin{definition} For any set $X$ and any functions $\beta_1:X\rightarrow
                                                                                                                K_1$,\ldots, $\beta_n:X\rightarrow K_n$, and any $T\in \Lambda^X_m$, we say that
                                                                                                                $\beta_1$,\ldots,$\beta_n$ are independent with respect to $T$ if for all
                                                                                                                $x_1,\ldots,x_m\in X$ and all $k_1\in K_1$, \ldots, $k_n\in K_n$,
                                                                                                                \begin{align*}\sum_{\substack{x\in \\\langle k_1 \rangle_{\!\beta_1}^{\phantom g}\cap\ldots\cap \langle k_n \langle k_n \rangle_{\!\beta_n}^{\phantom g}}}T(x|x_1,\ldots,
                                                                                                                x_m)=
                                                                                                                \bigg(\sum_{x\in \langle k_1 \rangle_{\!\beta_1}^{\phantom g}}T(x|x_1,\ldots,
                                                                                                                x_m)\bigg)\ldots\bigg(\sum_{x\in \langle k_n \rangle_{\!\beta_n}^{\phantom
                                                                                                                g}}T(x|x_1,\ldots, x_m)\bigg)\end{align*}
                                                                                                                \end{definition}
                                                                                                                \begin{proposition}For any set $X$ and any functions $\beta_1:X\rightarrow K_1$,\ldots,
                                                                                                                $\beta_n:X\rightarrow K_n$, and any $T\in \Lambda^X_m$, $\beta_1$,\ldots,$\beta_n$ are
                                                                                                                independent with respect to $T$ if and only if for all $x_1,\ldots,x_m\in X$ and all
                                                                                                                $k_1\in K_1$, \ldots, $k_n\in K_n$,

                                                                                                                \begin{multline*}\!\!\!\!\!\Xi_{\prod_{i=1}^n\beta_i}(T(\cdot\,|x_1,\ldots,x_m))((k_1,\ldots,k_n))=\\\Xi_{\beta_1}(T(\cdot\,|x_1,\ldots,x_m))(k_1)
                                                                                                                \ldots\Xi_{\beta_n}(T(\cdot\,|x_1,\ldots,x_m))(k_n)\end{multline*}
                                                                                                                \end{proposition}
                                                                                                                \ifthenelse{\boolean{showProofs}}{
                                                                                                                {\sc Proof:  }This proposition follows directly from the definition of
                                                                                                                the projection operator \defn{ProjectionOperator} and from the observation that \[\langle
                                                                                                                (k_1,\ldots,k_n)\rangle_{\prod_{i=1}^n\beta_i}=\langle
                                                                                                                k_1\rangle_{\beta_1}\cap,\ldots,\cap \langle k_n\rangle_{\beta_n}\quad\quad\qed\]}{}

                                                                                                                The next lemma shows that given some transmission function $T$ and some theme maps
                                                                                                                $\beta_1,\ldots,\beta_n$ that are independent with respect to $T$, if $T$ is ambivalent
                                                                                                                under each of the theme maps, then $T$ will also be ambivalent under the cartesian
                                                                                                                product of the theme maps.

                                                                                                                \begin{lemma} \label{CrossProductAmbivalent}For any $n\in \mathbb N^+$ let $\beta_1:X\rightarrow K_1$,\ldots,
                                                                                                                $\beta_n:X\rightarrow K_n$ be functions which share the same domain. For any $m\in
                                                                                                                \mathbb N^+$ let $T\in \Lambda^X_m$ be a transmission function such that
                                                                                                                $\beta_1,\ldots,\beta_n$ are independent with respect to $T$ and such that for all
                                                                                                                $i\in\{1,\ldots,n\}$, $T$ is ambivalent under $\beta_i$. Then $T$ is ambivalent under
                                                                                                                $\prod_{i=1}^n\beta_i$ with $\prod_{i=1}^n\beta_i$-projection
                                                                                                                $T^{\overrightarrow{\prod_{i=1}^n\beta_i}}$ given as follows,
                                                                                                                \begin{multline*}\!\!\!\!\!\!\!T^{\overrightarrow{\prod_{i=1}^n\beta_i}}((k_1,\ldots
                                                                                                                k_n)|(k^1_1,\ldots,k^n_1),\ldots,(k^1_m,\ldots,k^n_m))=\\T^{\overrightarrow{\beta_1}}(k_1|k^1_1,\ldots,k_m^1)
                                                                                                                \ldots T^{\overrightarrow{\beta_n}}(k_n|k^n_1,\ldots,k_m^n)\end{multline*}
                                                                                                                \end{lemma}\ifthenelse{\boolean{showProofs}}{{\sc Proof:  }
                                                                                                                 For all $x_1\in\langle (k_1^1,\ldots,k^n_1)\rangle_{\prod_{i=1}^n\beta_i}^{\phantom g},\ldots, x_m\in\langle
                                                                                                                (k_m^1,\ldots,k^n_m)\rangle_{\prod_{i=1}^n\beta_i}$
                                                                                                                \begin{align*}&T^{\overrightarrow{\prod_{i=1}^n\beta_i}}((k_1,\ldots
                                                                                                                k_n)|(k^1_1,\ldots,k^n_1),\ldots,(k^1_m,\ldots,k^n_m))\\&=
                                                                                                                \sum_{\substack{x\in\\\langle(k_1,\ldots,k_n)\rangle_{\prod_{i=1}^n\beta_i}}}T(x|x_1,\ldots,x_m)\\&=
                                                                                                                \sum_{\substack{x\in\\\langle k_1\rangle_{\!\beta_1}^{\phantom g}\cap,\ldots\cap\langle
                                                                                                                k_n\rangle_{\!\beta_n}^{\phantom g}}}T(x|x_1,\ldots,x_m)\end{align*} As
                                                                                                                $\beta_1,\ldots,\beta_n$ are independent with respect to $T$
                                                                                                                \begin{align*}&T^{\overrightarrow{\prod_{i=1}^n\beta_i}}((k_1,\ldots
                                                                                                                k_n)|(k^1_1,\ldots,k^n_1),\ldots,(k^1_m,\ldots,k^n_m))\\
                                                                                                                &=\bigg(\sum_{x\in \langle k_1 \rangle}T(x|x_1,\ldots,
                                                                                                                x_m)\bigg)\ldots\bigg(\sum_{x\in \langle k_n \rangle}T(x|x_1,\ldots,
                                                                                                                x_m)\bigg)\\
                                                                                                                &=T^{\overrightarrow{\beta_1}}(k_1|k^1_1,\ldots,k_m^1) \ldots
                                                                                                                T^{\overrightarrow{\beta_n}}(k_n|k^n_1,\ldots,k_m^n)\quad\quad\qed\phantom{aaaaaaaaaaaaaaa}\end{align*}
                                                                                                                }{}

                                                                                                                \section{On the Ambivalence of Common Variational Operators Used in SGAs}\label{simVarConsts}
                                                                                                                The cartesian product structure of the genome set and the nature of the variation
                                                                                                                operations commonly used in GAs makes it possible to obtain ambivalence results for transmission functions that model any of the common variation operators of an SGA.

                                                                                                                The following is some useful notation for dealing with SGAs. For any $n\in \mathbb Z^+$, let $\mathfrak B_n$ be the set of all bitstrings of length $n$. For any
                                                                                                                $x\in\mathfrak B_n$, and any $i\in\{1,\ldots,n\}$, let $x_i$ denote the $i^{th}$
                                                                                                                locus of $x$.

                                                                                                                Let $\ell\in\mathbb N^+$. For any $i\in \mathbb N^+$ such that $i\leq \ell$ let
                                                                                                                $\xi_i:\mathfrak B_\ell\rightarrow\{0,1\}$ be a theme map such that
                                                                                                                $\xi_i(x)=x_i$, i.e. $\xi_i$ maps a bitstring to the value of its $i^{th}$ locus.

                                                                                                                \begin{definition} \label{defSchemaMap}Let $I=\{j_1,\ldots,j_{|I|}\}$ be a subset of  $\{1,\ldots,\ell\}$,
                                                                                                                such that  $j_1<\ldots<j_{|I|}$. Let $\xi_{I}:\mathfrak
                                                                                                                B_\ell\rightarrow\mathfrak B_{|I|}$  denote the theme map
                                                                                                                $\xi_{j_1}\times\ldots\times\xi_{j_{|I|}}$.  We call such a theme map a \emph{schema map}
                                                                                                                \end{definition}
                                                                                                                Thus a schema map, maps any element of some schema $h$ to the bitstring that is obtained when all the wildcards are stripped out of the schema template of $h$.
                                                                                                                \subsection{Ambivalence of Canonical Mutation} Let us call a mutation operation that flips
                                                                                                                each bit independently with some fixed probability $\alpha$ \emph{canonical mutation} and
                                                                                                                let $M\in\Lambda^{\mathfrak B_\ell}_1$ be a transmission function that models this
                                                                                                                operation. Observe that for any $i\in\{1,\ldots,\ell\}$, $M$ is ambivalent under $\xi_i$
                                                                                                                with $M^{\overrightarrow{\xi_i}}\in\Lambda^{\mathfrak B_1}_1$ given as follows:
                                                                                                                \[M^{\overrightarrow{\xi_i}}(k|\,l)=\bigg\{\begin{array}{cl}\alpha &\text{ if
                                                                                                                $k\not=l$}\\1-\alpha & \text{ otherwise}\end{array}\]

                                                                                                                Observe that for any $I\in \mathbb I$, such that $I=\{j_1,\ldots,j_{|I|}\}$ and
                                                                                                                $j_1<\ldots<j_{|I|}$, $\xi_{j_1},\ldots,\xi_{j_{|I|}}$ are independent with respect
                                                                                                                to $M$ as $M$ models a mutation operator that flips bits \emph{independently} of each
                                                                                                                other. Hence, by lemma \ref{CrossProductAmbivalent}, $M$ is ambivalent under $\xi_I$
                                                                                                                with $M^{\overrightarrow{\xi_I}}\in\Lambda^{\mathfrak B_{|I|}}_1$ given by
                                                                                                                \[ M^{\overrightarrow{\xi_I}}(k|\,l)=M^{\overrightarrow{\xi_{j_1}}}(k_1|\,l_1)\ldots
                                                                                                                M^{\overrightarrow{\xi_{j_{|I|}}}}(k_{|I|}\,|\,l_{|I|}).1\] Multiplication by $1$ at
                                                                                                                the end of the right hand side of this expression is necessary to account for the case
                                                                                                                when $I$ is the empty set.
                                                                                                                \subsection{Ambivalence of Mask Based Crossover}
                                                                                                                Let $\Psi$ be the set of all masks for length $\ell$ bitstrings, i.e. $\mathfrak B_\ell$ is itself a set
                                                                                                                of bitstrings in $\mathfrak B_\ell$. For any mask $\psi\in \mathfrak B_\ell$ let the $2$-parent
                                                                                                                transmission function $T_\psi\in\Lambda^{\mathfrak B_\ell}_2$ be defined as follows:\\
                                                                                                                \[T_\psi(x|y,z)=\text{{\LARGE$\bigg\{$}}\begin{array}{cl}1&\text{ if $\forall i\in\{1,\ldots,\ell,\},$}
                                                                                                                \\ &\indent(x_i=y_i
                                                                                                                \wedge \psi_i=0) \,\,\,\vee\\ &\indent\,\,\,\,\,\,(x_i=z_i \wedge \psi_i=1)\\
                                                                                                                0&\text{ otherwise}\end{array}\]

                                                                                                                Note that for any parents $y,z$ $T_\psi(\cdot|y,z)$ is a discrete delta function which
                                                                                                                concentrates all its distribution mass on one child. In other words, $T_\psi$ is
                                                                                                                `deterministic'.

                                                                                                                For all $\psi$ and for all $i\in\{1,\ldots,\ell\}$, see that $T_\psi$ is ambivalent under
                                                                                                                $\xi_i$ with $T_\psi^{\overrightarrow{\xi_i}}\in\Lambda^{\{0,1\}}_2$ given as
                                                                                                                follows:
                                                                                                                \[T_\psi^{\overrightarrow{\xi_i}}(k|\,l,m)=\text{{\Large$\bigg\{$}}\begin{array}{cl}1&\text{
                                                                                                                if }(k=l
                                                                                                                \wedge \psi_i=0) \,\,\,\vee\\ &\indent\,\,\,\,\,\,(k=m \wedge \psi_i=1)\\
                                                                                                                0&\text{ otherwise}\end{array}\]

                                                                                                                Observe that for any $I\in \mathbb I$, such that $I=\{a_1,\ldots,a_{|I|}\}$ and
                                                                                                                $a_1<\ldots<a_{|I|}$, $\xi_{a_1},\ldots,\xi_{a_{|I|}}$ are independent with respect
                                                                                                                to $T_\psi$ because for any two parents $y,z$ the mutual information between the
                                                                                                                distributions $\Xi_{\xi_{a_1}} (T_\psi(\cdot|y,z)),\ldots,\Xi_{\xi_{a_{|I|}}}
                                                                                                                (T_\psi(\cdot|y,z))$ is zero. (This is because the distribution $\Xi_{\xi_i}
                                                                                                                (T_\psi(\cdot|y,z))$ depends only on the values $y_i, z_i$, and $\psi_i$). Hence, by
                                                                                                                lemma \ref{CrossProductAmbivalent}, $T_\psi$ is ambivalent under $\xi_I$ with
                                                                                                                $T_\psi^{\overrightarrow{\xi_I}}$ given as follows:
                                                                                                                \[ T^{\overrightarrow{\xi_I}}_\psi(k|\,l,m)=T^{\overrightarrow{\xi_{a_1}}}_\psi(k_1|\,l_1,m_1)\ldots
                                                                                                                T^{\overrightarrow{\xi_{a_{|I|}}}}_\psi(k_{|I|}\,|\,l_{|I|},m_{|I|}).1\] Multiplication
                                                                                                                by $1$ at the end of the right hand side of this expression is once again necessary to
                                                                                                                account for the case when $I$ is the empty set.

                                                                                                                For some choice of distribution over the set of all masks $q\in\Lambda^\mathfrak B_\ell$, let
                                                                                                                $T\in\Lambda^{\mathfrak B_\ell}_2$ be given as follows:
                                                                                                                \[T(x|y,z)=\sum_{\psi\in\mathfrak B_\ell}q(\psi)T_\psi(x|y,z)\]As stated above, for all $\psi\in \mathfrak B_\ell$,  $T_\psi$ is ambivalent under $\xi_I$.
                                                                                                                Therefore by lemma \ref{WeightedSumAmbivalent}, $T$ is ambivalent under $\xi_I$ with
                                                                                                                $T^{\overrightarrow{\xi_I}}\in\Lambda^{\mathfrak B_{|I|}}_2$ given
                                                                                                                by\[T^{\overrightarrow{\xi_I}}(k|\,l,m)=\sum_{\psi\in\mathfrak B_\ell}q(\psi)T_\psi^{\overrightarrow{\xi_I}}(k|\,l,m)\]
                                                                                                                By appropriately choosing $q$, $T$ can be made to model any $\ell$-point or uniform
                                                                                                                crossover operation \cite{conf/gecco/WrightVR03}, so any $\ell$-point or uniform crossover
                                                                                                                operation is ambivalent under $\xi_I$.

                                                                                                                Finally, by lemma \ref{CompositionAmbivalent} any composition of $n$-point or uniform
                                                                                                                crossover operations with the canonical mutation operation, in any order, is ambivalent
                                                                                                                under $\xi_I$.

\section{Sufficient Conditions for Coarse-Graining IPSGA Dynamics}\label{suffConds}

We now use the result in the previous section to argue that the dynamics of an IPSGA with long genomes, uniform crossover, and fitness proportional selection can be coarse-grained with high fidelity for a relatively coarse schema map, provided that the initial population satisfies a constraint called \emph{approximate schematic uniformity} and the fitness function satisfies a  constraint called \emph{low-variance schematic fitness distribution}. We stress at the outset that our argument is principled but informal, i.e. though the argument rests relatively straightforwardly on theorem 3, we do find it necessary in places to appeal to the reader's intuitive understanding of GA dynamics. In the following chapter we will experimentally validate our conclusions.

For any $n\in \mathbb Z^+$, let $\mathfrak B_n$ be the set of all bitstrings of length $n$. For some $\ell\gg1$ and some $m\ll \ell$, let $\beta:\mathfrak B_\ell\rightarrow \mathfrak B_m$ be some schema theme map. Let $f^*:\mathfrak B_m\rightarrow \mathbb R^+$ be some function. For each $k\in \mathfrak B_m$, let  $D_k\in\Lambda^{\mathbb R^+}$ be some distribution over the reals with low variance such that the mean of distribution $D_k$ is $f^*(k)$. Let $f:\mathfrak B_\ell\rightarrow \mathbb R^+$ be a fitness function such that for any $k\in \mathfrak B_m$, the fitness values of the elements of $\langle k \rangle_\beta$ are independently drawn from the distribution $D_k$. For such a fitness function we say that fitness is \emph{schematically distributed with low-variance}. We call the distributions $D_k$ \emph{schema fitness distributions}.

Let $U$ be a set of distributions such that for any $k\in \mathfrak B_m$ and any $p\in U$, $\mathcal C_\beta(p,k)$ is approximately uniform. It is easily checked that $U$ satisfies the condition $\Xi_\beta(U)=\Lambda^{\mathfrak B_m}$. We say that the distributions in $U$ are \emph{approximately schematically uniform}.

Let $\delta$ be the minimal bound such that for all $p\in U$ and for all $k\in\mathfrak B_m$, $ |\mathcal E_f\circ\mathcal C_\beta(p,k)-f^*(k)|\leq\delta$. Then,  for any $\epsilon>0$, $\mathbf P(\delta< \epsilon) \rightarrow 1$ as $\ell-m\rightarrow\infty$. Because we have chosen $\ell$ and $m$ such that $\ell-m$ is `large', it is reasonable to assume that the minimal bound on the schematic mean divergence of $f$ on $U$ under $\beta$ is likely to be `low'.

Let $T\in \Lambda^{\mathfrak B_\ell}$ be a transmission function that models the application of uniform crossover. In section \ref{simVarConsts} we proved that a transmission function that models any mask based crossover operation is ambivalent under any schema map. Uniform crossover is mask based, and $\beta$ is a schema map, therefore $T$ is ambivalent under $\beta$.

Let $p_{\frac{1}{2}}\in \Lambda^{\mathfrak B_1}$ be such that $p_{\frac{1}{2}}(0)=\frac{1}{2}$ and $p_{\frac{1}{2}}(1)=\frac{1}{2}$. For any $p \in U$, $\mathcal S_fp$ may be `outside' $U$ because there may be one or more $k\in \mathfrak B_m$ such that $\mathcal C_\beta(\mathcal S_fp,k)$ is not quite uniform. Recall that for any $k\in \mathfrak B_m$ the variance of $D_k$ is low. Therefore even though $\mathcal S_fp$ may be `outside' $U$, the deviation from schematic uniformity is not likely to be large. Furthermore, given the low variance of $D_k$, the marginal distributions of $\mathcal C_\beta(\mathcal S_fp,k)$ will be very close to $p_{\frac{1}{2}}$. Given these facts and our choice of transmission function, for all $k\in K$, $\mathcal C_\beta(\mathcal V_T\circ\mathcal S_fp,k)$ will be more uniform than $\mathcal C_\beta(\mathcal S_fp,k)$, and we can assume that $\mathcal V_T\circ\mathcal S_fp$ is in $U$. In other words, we can assume that $E$ is non-departing over $U$.

Let $E=(\mathfrak B_\ell, T, f)$ and $E^*=(\mathfrak B_m, T^{\overrightarrow\beta}, f^*)$ be evolution machines. By the discussion above and the limitwise coarsenablity of evolution theorem one can expect that for any approximately thematically uniform distribution $p\in U$ (including of course the uniform distribution over $\mathfrak B_\ell$), the dynamics of $E^*$ when initialized with $\Xi_\beta p$ will approximate the projected dynamics of $E$ when initialized with $p$. As the bound $\delta$ is `low', the fidelity of the approximation will be `high'.

Note that the constraint that fitness be low-variance schematically distributed, which is required for this coarse-graining, is much weaker than the very strong constraint of schematic fitness invariance (all genomes in each schema must have the \emph{same} value) which \citet{conf/gecco/WrightVR03} argue is required to coarse-grain IPSGA dynamics.

\section{Experimental Validation}\label{expval}
Let $\mathcal F$ be some family of schemata of order $o$ and length $\ell$, and let $\xi:\mathcal F\rightarrow\mathfrak B_{o}$ be the bijection that maps any schema in $\mathcal F$ to the bitstring that is obtained when all the wildcards are removed from that schema. Consider an IPSGA with genome set $\mathfrak B_\ell$, fitness proportional selection, and uniform crossover. Let its fitness function be such that  fitness is low-variance schematically distributed with respect to $\mathcal F$ and let the function $\psi:\mathcal F\rightarrow \Lambda^{\mathbb R^+}$ be a mapping between the schemata in $\mathcal F$ and their corresponding low-variance fitness distributions. Let the initial population be uniformly distributed over the genome set. We shall refer to this infinite population IPSGA as \textsc{ipsga1}. A direct experimental validation of the conclusions of section \ref{suffConds} would show that the exact dynamics (i.e. frequencies over multiple generations) of the schemata in $\mathcal F$ under the action of  \textsc{ipsga1} can be approximated by the coarsened dynamics of this IPSGA, where the theme map used in the coarsening is a schema theme map which maps the genomes in $\mathfrak B_\ell$ to the schemata in $\mathcal F$. As per the conclusions in section \ref{suffConds} these coarsened dynamics are given by the dynamics of an IPSGA with genome set $\mathfrak B_o$, fitness proportional selection, uniform crossover, uniformly distributed initial population, and fitness function $\psi\circ\xi^{-1}$. We shall refer to this IPSGA as \textsc{ipsga2}.

As described in section \ref{TowPrincTheory} an exact calculation of the dynamics of the schemata in $\mathcal F$ under \textsc{ipsga1} has time complexity that is exponential in $\ell$. Therefore such a calculation quickly becomes infeasible as $\ell$ gets large. However, because the conclusions in section \ref{suffConds} are premised upon $o$ being much smaller than $\ell$, in order to experimentally validate these conclusions $\ell$ needs to be large.

\subsection{An Indirect But Computationally Feasible Approach to Experimental Validation} \label{indirectExperimentalValidation}
We  resolve this dilemma by making a key assumption and a particular modeling decision. Consider an SGA with genomes of length $\ell$, fitness proportional selection, uniform crossover, a finite population of size $N\gg 2^o$ and the same fitness function as the IPGA described above. We will assume that the dynamics of the schemata in $\mathcal F$ under the action of this SGA approximates the dynamics of same family of schemata under the action of the IPSGA. Tracking these dynamics in the former case is exponential in $o$ and \emph{linear} in $\ell$, and is therefore feasible when  $\ell$ is large, provided that  $o$ is small.

 For genomes of any significant length, predetermining the fitness values of all possible genomes that may be generated by the SGA is computationally infeasible. So instead what if each time a genome is generated the SGA determines its fitness value ``on the fly" by drawing a random sample from the distribution corresponding to the the (unique) schema in $\mathcal F$ that the genome belongs to? There is a problem with this approach. If a genome is generated more than once during the course of a run it is unlikely to be assigned the same fitness value each time it gets generated. One way to deal with this problem is by storing all generated genomes and their fitness values and using this information to ensure that each genome is always assigned a unique fitness value by the fitness function. A second, far simpler approach, is based on the following observation: as  $\ell$ gets larger the probability that some genome will be generated more than once during an evolutionary run gets smaller. Therefore by letting $\ell$ be ``large enough" we can assume that the same genome is never generated more than once during some finite number of evolutionary runs (provided that the number of generations in each run is also finite). In other words, by letting $\ell$ be ``very large" we are free to determine the fitness value of each generated genome ``on the fly" and do not need to  maintain a store of previously generated genomes and their fitness values.

\begin{figure}[t]\begin{center}
\includegraphics[height=10cm, width=\textwidth]{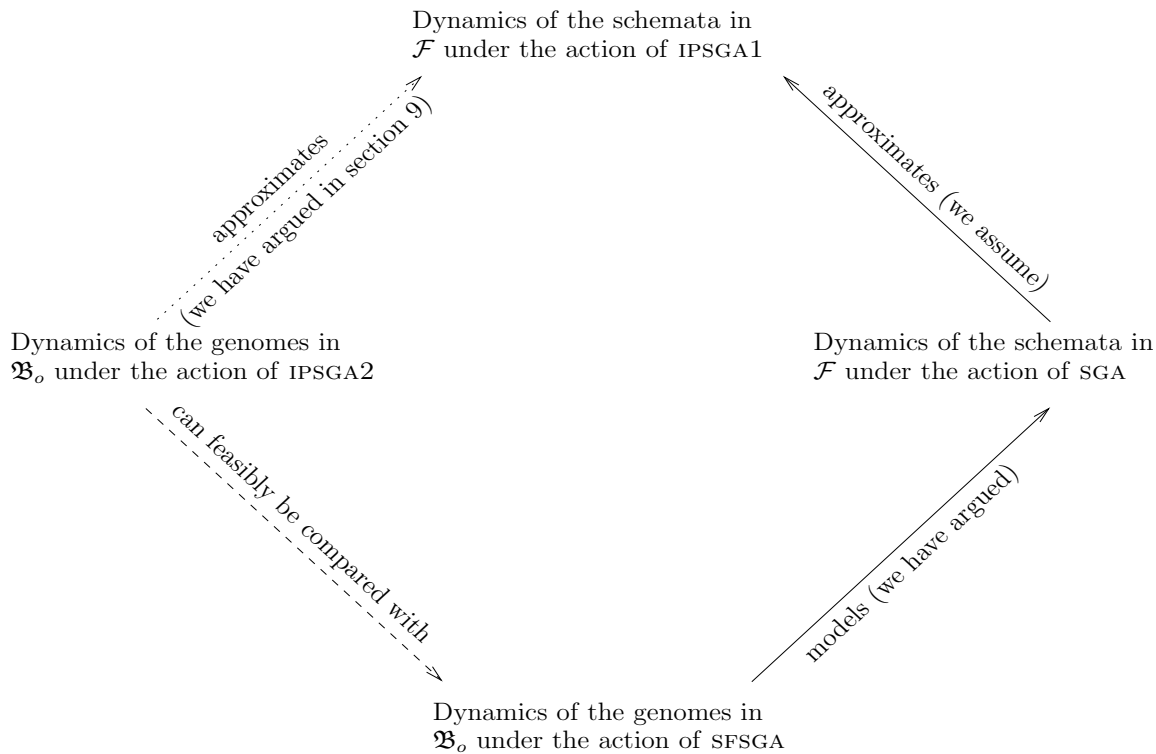}\end{center}
\caption{A diagrammatic depiction of the assumption and modeling decision which allows us to indirectly compare the actual frequency dynamics of the schemata of length $\ell$ and order $o$ in a schema family $\mathcal F$ under the action of  \textsc{ipsga1} with the frequency dynamics of these schemata predicted by a coarsening  of the genomic frequency dynamics of \textsc{ipsga1} which uses a schema map from $\mathfrak B_\ell$ to $\mathfrak B_o$ and partitions $\mathfrak B_\ell$ into the schemata in $\mathcal F$.}  \label{coarsegrainingValidation}\end{figure}

Even though the time complexity of the SGA described above is linear in $\ell$, for large enough $\ell$ this SGA becomes infeasible to execute. So a question that seems to require a precise answer is ``how large is `very large'?" We will now argue that a precise answer to this question is not required. Let us call the SGA described in the previous paragraph \textsc{sga}.  Let \textsc{sfsga} be an SGA with population size $N$, genome set $\mathfrak B_{o}$, fitness proportional selection, uniform crossover, and a \emph{stochastic} fitness function, which for any genome $g\in\mathfrak B_{o}$ returns a sample from the distribution $\xi\circ\psi^{-1}(g)$. Then with a little thought the reader can see that the frequency dynamics, under the action of \textsc{sga}, of any schema $k\in \mathcal F$ can is modeled by the frequency dynamics, under the action of \textsc{sfsga}, of the genome $\xi(k)$. Since we can use $\textsc{sfsga}$ to model the frequency dynamics of the schemata in $\mathcal F$ under the action of $\textsc{sga}$ the answer to the question ``how large does $\ell$ need to be" is simply ``large enough that it is unlikely that any genome in $\mathfrak B_\ell$ is generated more than once during the course of an experiment".

The dynamics of \textsc{ipsga2} can feasibly be calculated when $o$ is small (regardless of the length of $\ell$). Therefore when $o$ is small it is possible to compare the coarsened dynamics of the schemata in $\mathcal F$ with an approximation of these dynamics obtained by using \textsc{sfsga}. The diagram in Figure \ref{coarsegrainingValidation} displays the scheme described above in graphical format.
\subsection{Experiments}
For a number of different parameter regimes we now compare the dynamics of the genomes in $\mathfrak B_o$ under the action of \textsc{ipsga2} with the dynamics of the genomes in $\mathfrak B_o$ under the action of \textsc{sfsga}. In each experiment each genome $g\in \mathfrak B_o$ is associated with some value $f_g$ chosen from the interval [2,3]. We call this value the f-value of that genome. The stochastic fitness function of \textsc{sfsga} is such that whenever $g$ is generated it assigns to $g$ a fitness value sampled from the distribution $\mathcal N(f_g,\sigma^2)$ (where $\mathcal N(m,n)$ denotes the normal distribution with mean $m$ and variance $n$). If the sample is less then 0 then the stochastic fitness function returns the value 0. The standard deviation $\sigma$ is 0.8 in all our experiments. The fitness function of \textsc{ipsga2} simply maps the genome $g$ to the value $f_g$. The maximum number of generations in all experiments is 30. Each experiment consists of a single run of \textsc{ipsga2} and $r$ runs of \textsc{sfsga}.  For each genome, the frequencies of that genome in each generation across all runs is averaged and plotted. For values of $r$ greater than 1, the error bars in the plots give the standard deviation of the frequency of the  genome in each generation. In each experiment the size of the population of \textsc{sfsga} is denoted by $N$.

\subsubsection{Experiment 1}
\begin{figure}
\includegraphics[height=16cm, width=\textwidth]{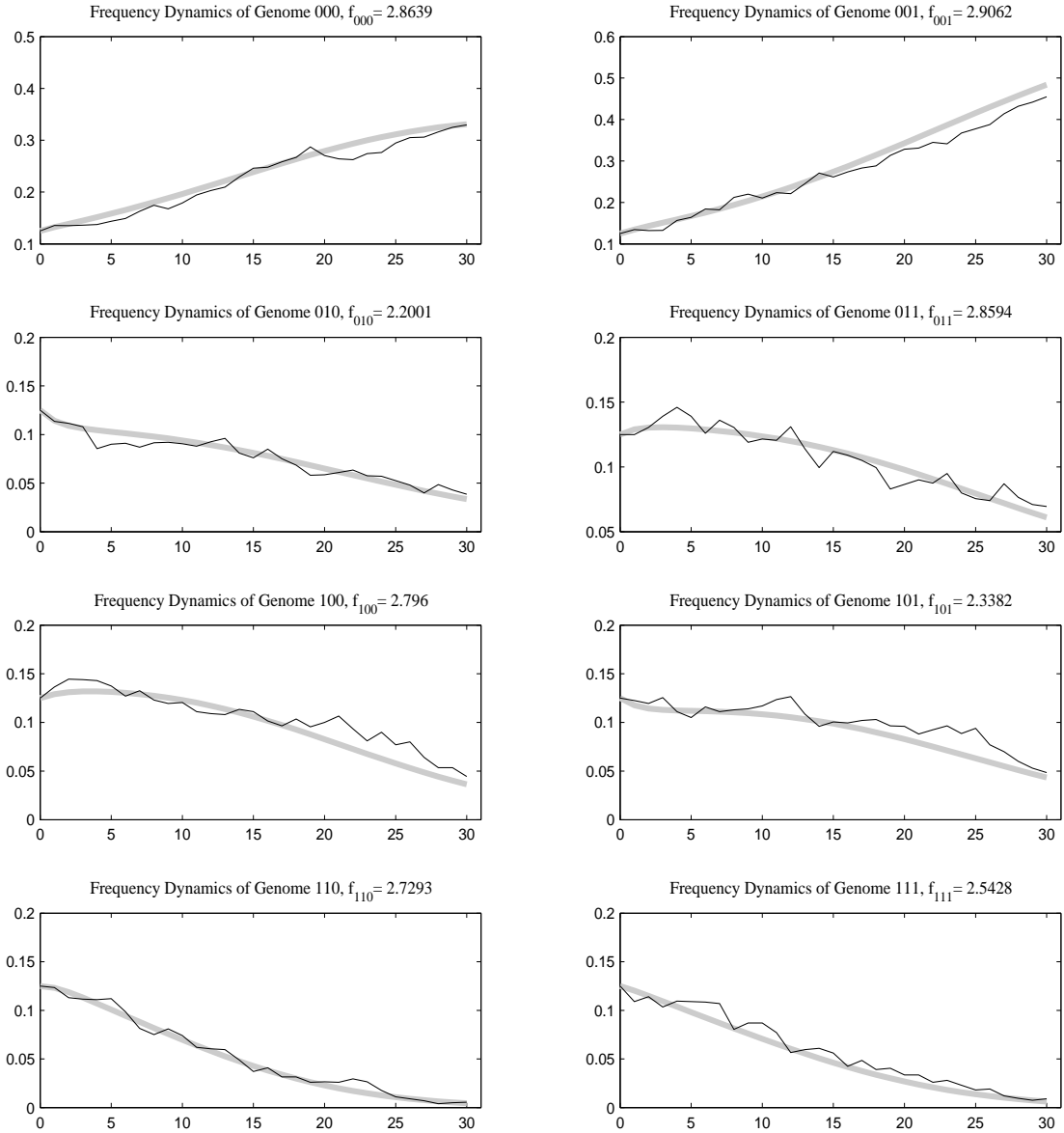}
\caption{Results of Experiment 1 ($o=3$, $N=2000$, $r=1$). A series of eight  plots showing the dynamics of the eight genomes in $\mathfrak B_3$ under the action of \textsc{ipsga2} and \textsc{sfsga}. The independent axis in each plot shows the generation number, and the dependent axis gives the frequency of  a genome in a population. The title of each plot displays the genome and its f-value. In each plot the thick light grey line shows the frequency dynamics of the genome under the action of  \textsc{ipsga2}. The thin black line shows the frequency dynamics of the genome under the action of \textsc{sfsga}}\label{exp1}
\end{figure}

In the first experiment  $o=3$, $N=2000$, and $r=1$. Figure \ref{exp1} shows the (frequency) dynamics of the genomes in $\mathfrak B_3$ under the action of \textsc{ipsga2} and \textsc{sfsga}.
\subsubsection{Experiment 2}
\begin{figure}
\includegraphics[height=16cm, width=\textwidth]{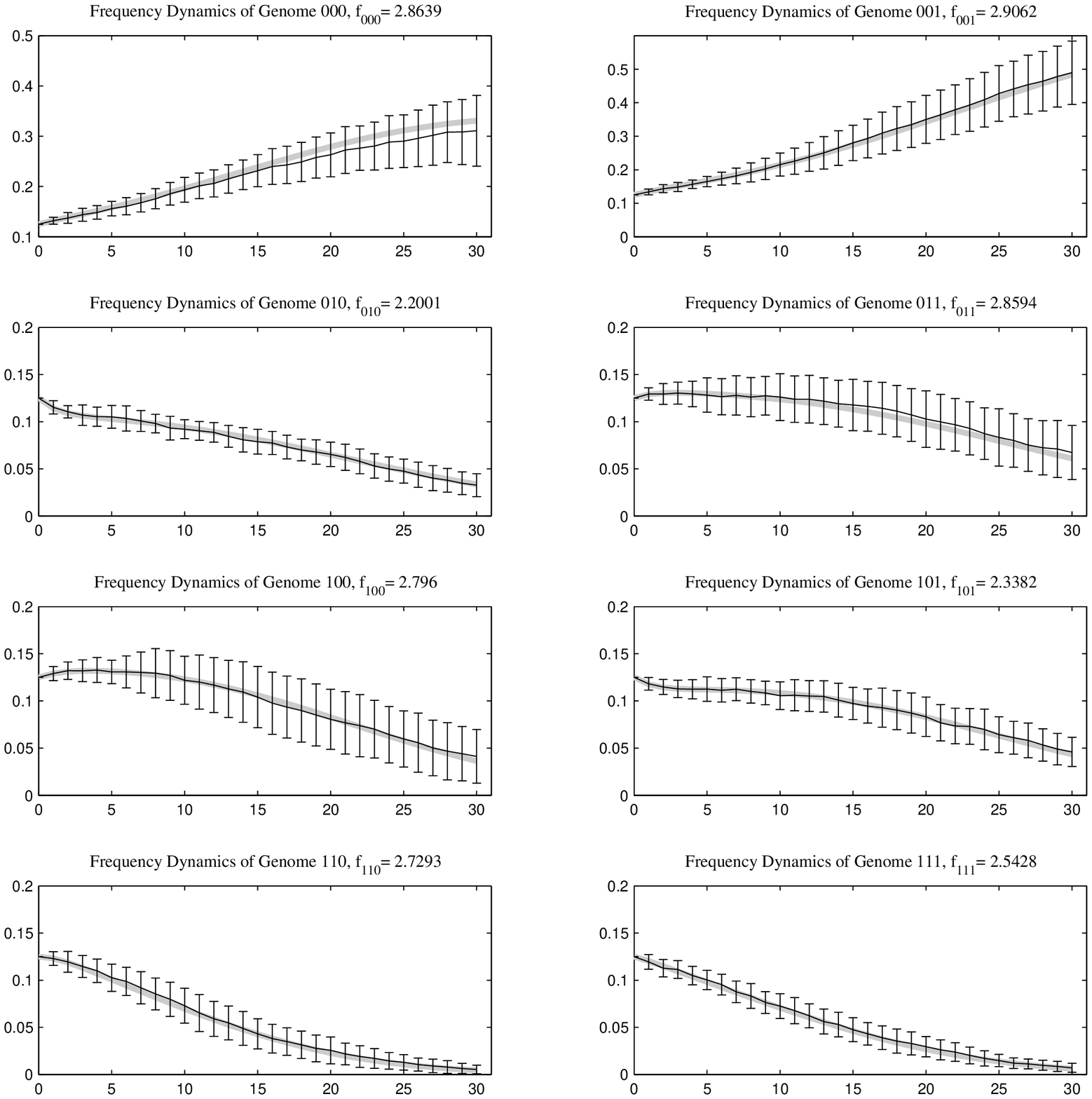}
\caption{Results of Experiment 2 ($o=3$, $N=2000$, $r=40$). A series of eight  plots showing the dynamics of the eight genomes in $\mathfrak B_3$ under the action of \textsc{ipsga2} and \textsc{sfsga}. The independent axis in each plot shows the generation number, and the dependent axis gives the frequency of  a genome in a population. The title of each plot displays the genome and its f-value. In each plot the thick light grey line shows the frequency dynamics of the genome under the action of  \textsc{ipsga2}. The thin black line shows the average frequency dynamics of the genome under the action of \textsc{sfsga} over 40 runs. The error bars show one standard deviation above and below the average in each generation.}\label{exp2}
\end{figure}

The next experiment builds confidence that the similarity between the dynamics of the genomes under the action of \textsc{sfsga} and the dynamics of the genomes under the action of \textsc{ipsga2} in experiment 1 was no fluke. Figure \ref{exp2} shows the result of this experiment in which we raise the value of  $r$ to $40$.
\subsubsection{Experiment 3}
\begin{figure}
\includegraphics[height=16cm, width=\textwidth]{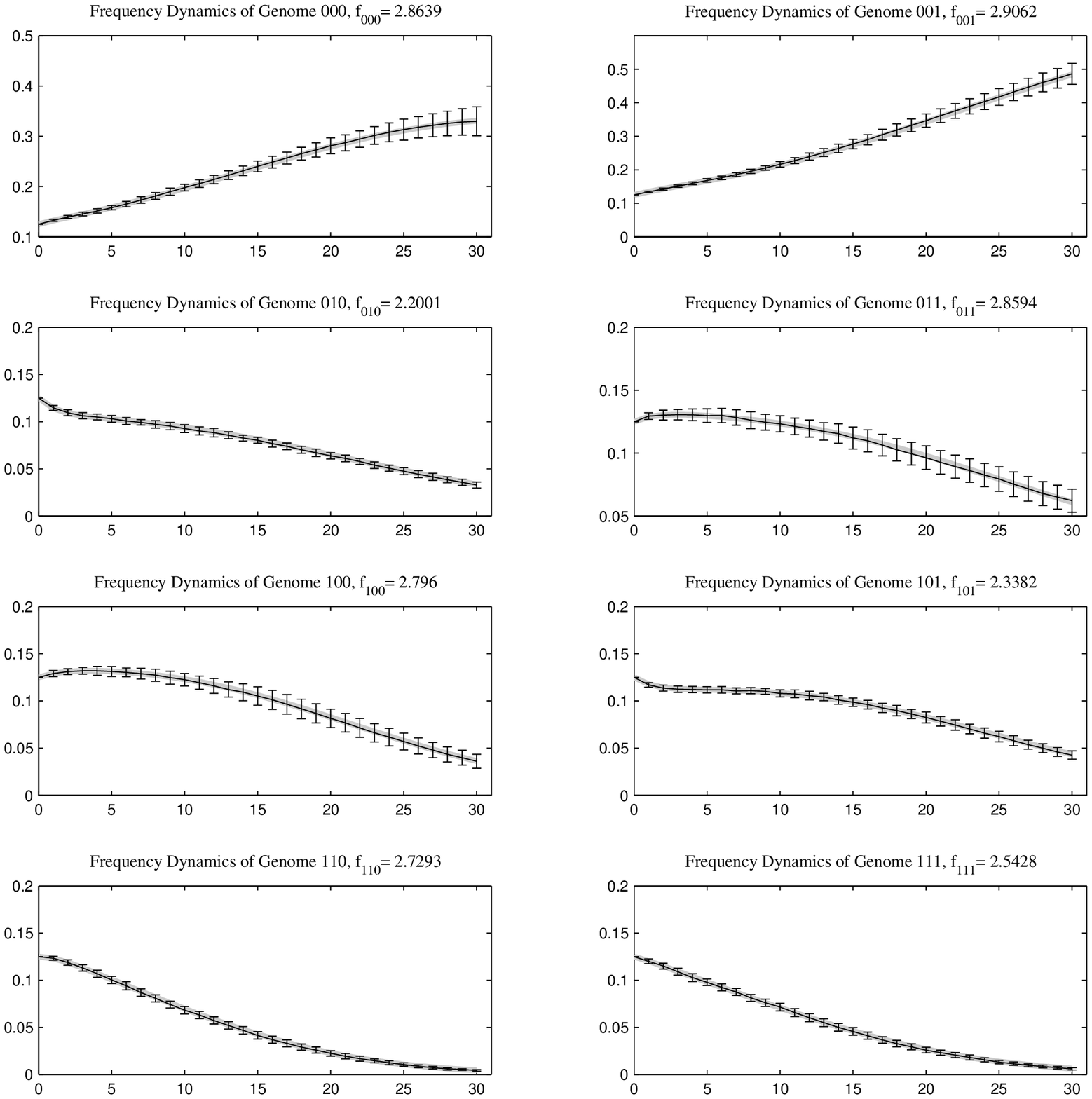}
\caption{Results of Experiment 3 ($o=3$, $N=20000$, $r=40$). A series of eight  plots showing the dynamics of the eight genomes in $\mathfrak B_3$ under the action of \textsc{ipsga2} and \textsc{sfsga}. The independent axis in each plot shows the generation number, and the dependent axis gives the frequency of  a genome in a population. The title of each plot displays the genome and its f-value. In each plot the thick light grey line shows the frequency dynamics of the genome under the action of  \textsc{ipsga2}. The thin black line shows the average frequency dynamics of the genome under the action of \textsc{sfsga} over 40 runs. The error bars show one standard deviation above and below the average in each generation.}\label{exp3}
\end{figure}

Figure \ref{exp3}  shows the results of our fourth experiment in which we increase the value of $N$ to $20000$.  Note how the error bars are much smaller than those in experiment 3..
 \subsubsection{Experiment 4}
\begin{figure}
\includegraphics[height=16cm, width=\textwidth]{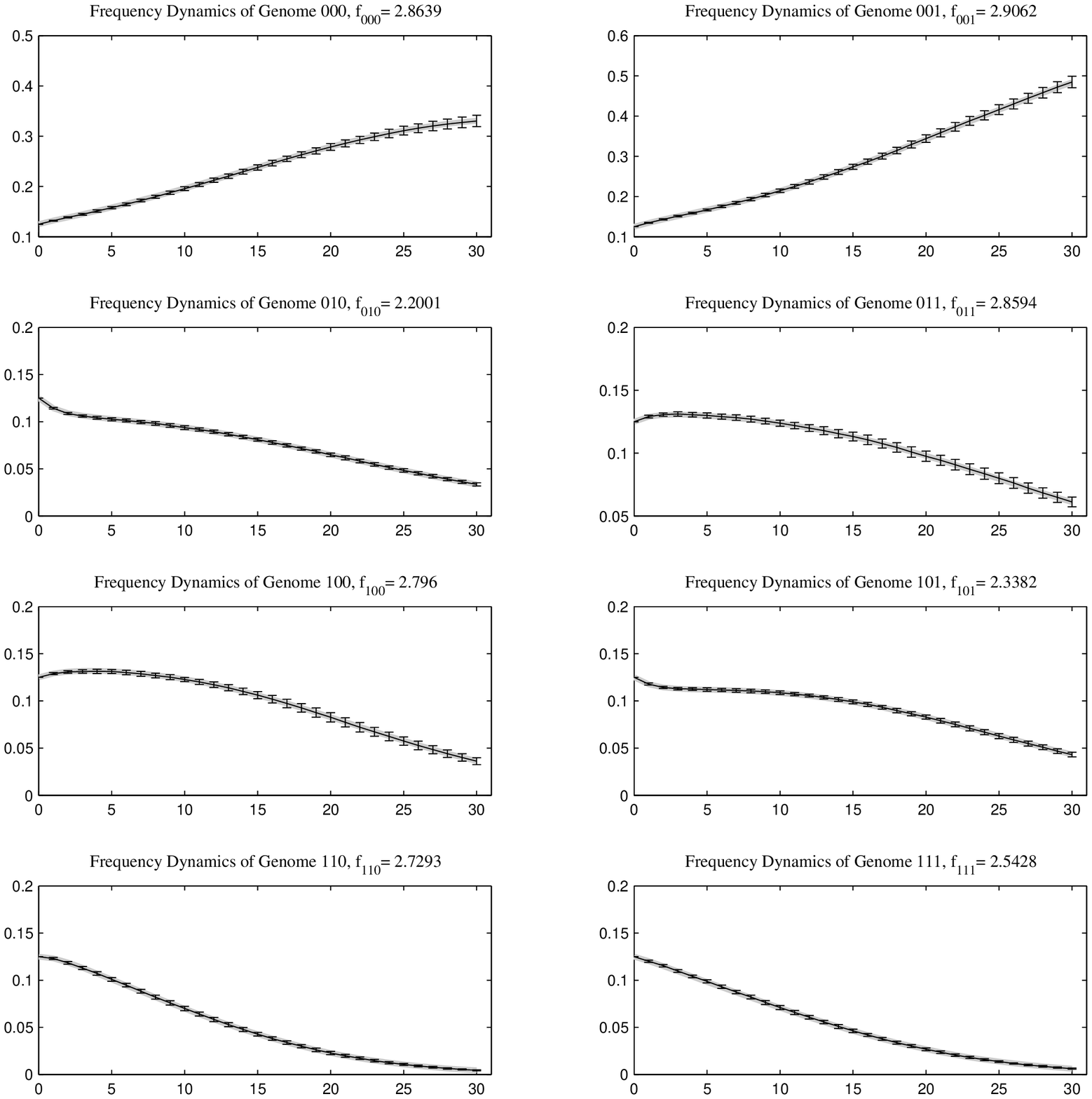}
\caption{Results of Experiment 4 ($o=3$, $N=100000$, $r=40$). A series of eight  plots showing the dynamics of the eight genomes in $\mathfrak B_3$ under the action of \textsc{ipsga2} and \textsc{sfsga}. The independent axis in each plot shows the generation number, and the dependent axis gives the frequency of  a genome in a population. The title of each plot displays the genome and its f-value. In each plot the thick light grey line shows the frequency dynamics of the genome under the action of  \textsc{ipsga2}. The thin black line shows the average frequency dynamics of the genome under the action of \textsc{sfsga} over 40 runs. The error bars show one standard deviation above and below the average in each generation.}\label{exp4}
\end{figure}

The error bars are smaller still in figure \ref{exp4} in which the consequence of further increasing the value of $N$ to $100000$ can be seen. From this experiment and the previous two, one can infer that as the population size of the \textsc{sfsga} gets larger the dynamics of the genomes under the action of \textsc{sfsga}more closely mirrors the dynamics of the genomes under the action of \textsc{ipsga2}
\subsubsection{Experiment 5}
\begin{figure}
\includegraphics[height=16cm, width=\textwidth]{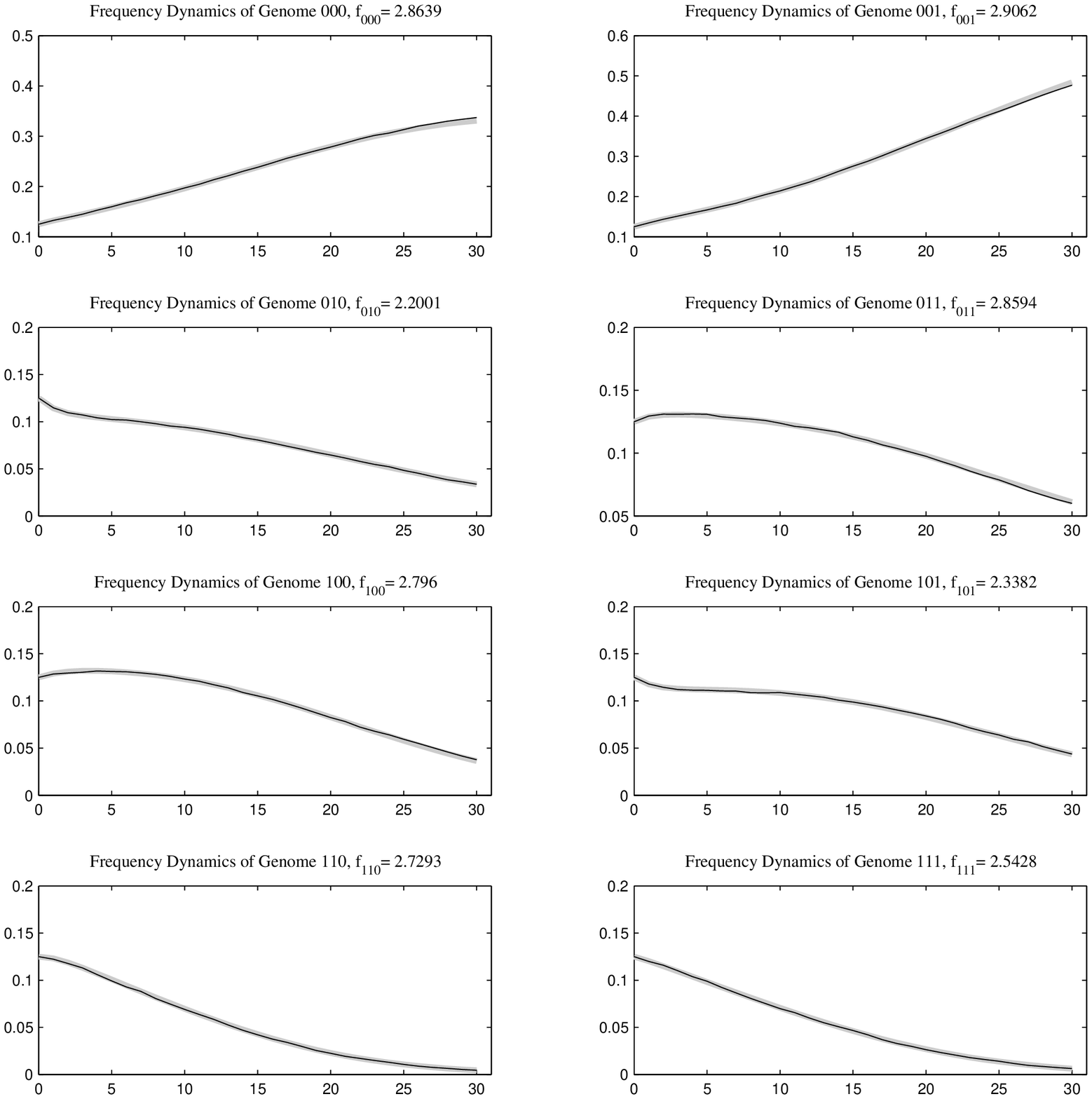}
\caption{Results of Experiment 5 ($o=3$, $N=400000$, $r=1$). A series of eight  plots showing the dynamics of the eight genomes in $\mathfrak B_3$ under the action of \textsc{ipsga2} and \textsc{sfsga}. The independent axis in each plot shows the generation number, and the dependent axis gives the frequency of  a genome in a population. The title of each plot displays the genome and its f-value. In each plot the thick light grey line shows the frequency dynamics of the genome under the action of  \textsc{ipsga2}. The thin black line shows the frequency dynamics of the genome under the action of \textsc{sfsga}}\label{exp5}
\end{figure}

Experiment 5 validates this inference. We set $r$ back to 1, and set $N$ to $400000$. Figure \ref{exp5} shows the result of this experiment. Note the close match between the dynamics of the genomes under the action of \textsc{ipsga2} and \textsc{sfsga}
\subsubsection{Experiment 6}
\begin{figure}
\includegraphics[height=16cm, width=\textwidth]{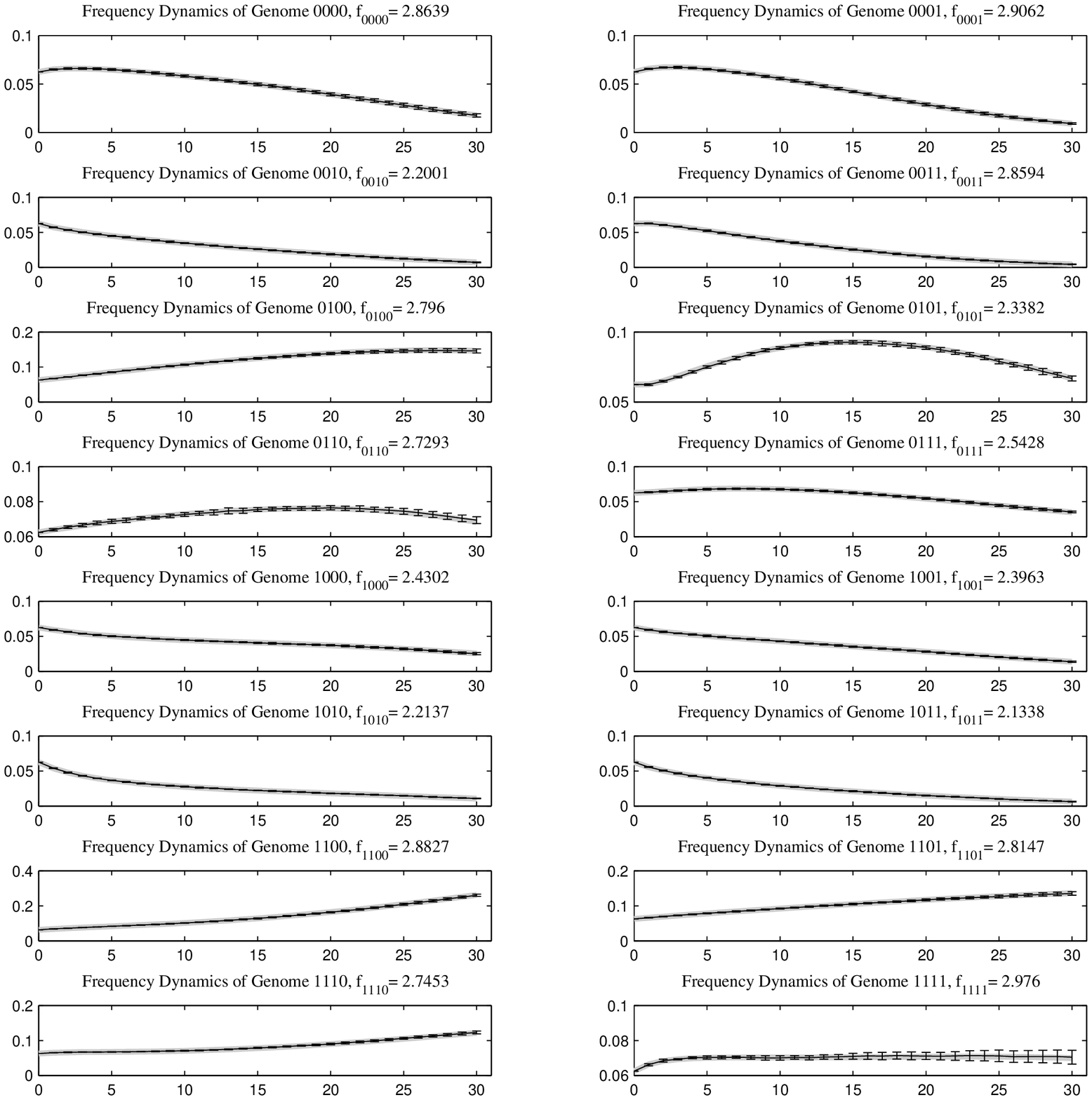}
\caption{Results of Experiment 6 ($o=4$, $N=200000$, $r=10$). A series of sixteen plots showing the dynamics of the sixteen genomes in $\mathfrak B_4$ under the action of \textsc{ipsga2} and \textsc{sfsga}. The independent axis in each plot shows the generation number, and the dependent axis gives the frequency of  a genome in a population. The title of each plot displays the genome and its f-value. In each plot the thick light grey line shows the frequency dynamics of the genome under the action of  \textsc{ipsga2}. The thin black line shows the average frequency dynamics of the genome under the action of \textsc{sfsga} over 10 runs. The error bars show one standard deviation above and below the average in each generation.}\label{exp6}
\end{figure}

In the final experiment we show that there is nothing special about $o=3$. Figure \ref{exp6} shows the results of an experiment in which  $o=4$, $N=200000$ and $r=10$.

\section{Rescuing the SGA From a Perceived Limitation} \label{rescuing}
We can use the model in the previous section (\textsc{sfsga}) to debunk a common misconception about SGAs which, in our opinion, will be retrospectively judged to be a significant barrier to the discovery of an accurate theory of adaptation for the SGA. It is widely assumed that SGAs are incapable of increasing in the frequency of  a low-order schema with above average fitness when the defining bits of that schema are widely dispersed \citep{goldberg:1989:mgamafr,desInnov}. We now argue that this assumption is false. Our argument is independent of the theoretical work in sections \ref{MathematicalPreliminaries}--\ref{suffConds} however it does rely on the uncontroversial modeling decision that we made in section \ref{expval}.

Consider an SGA with long genomes. For a concrete example let us say that the genomes are of length $2\times10^9+1$. Suppose that selection is fitness proportional, crossover is uniform, the population size of the SGA is $1000$, the the initial population is always drawn from a uniform distribution over the genome space, and fitness is low-variance schematically distributed with respect to some family of schemata $\mathcal F$ of order 3. Suppose the defining bits of the schemata are in positions $1, 10^9+1,$ and $2\times10^9+1$.  Let this SGA play the role of \textsc{sga} in the discussion in section \ref{expval}. Then as described in section \ref{expval} the the frequency dynamics of the schemata in $\mathcal F$ under the action of this SGA can be modeled by the frequency dynamics of the genomes in $\mathfrak B_3$ under the action of an SGA with a stochastic fitness function \textsc{sfsga}.

\begin{figure}
\includegraphics[height=16cm, width=\textwidth]{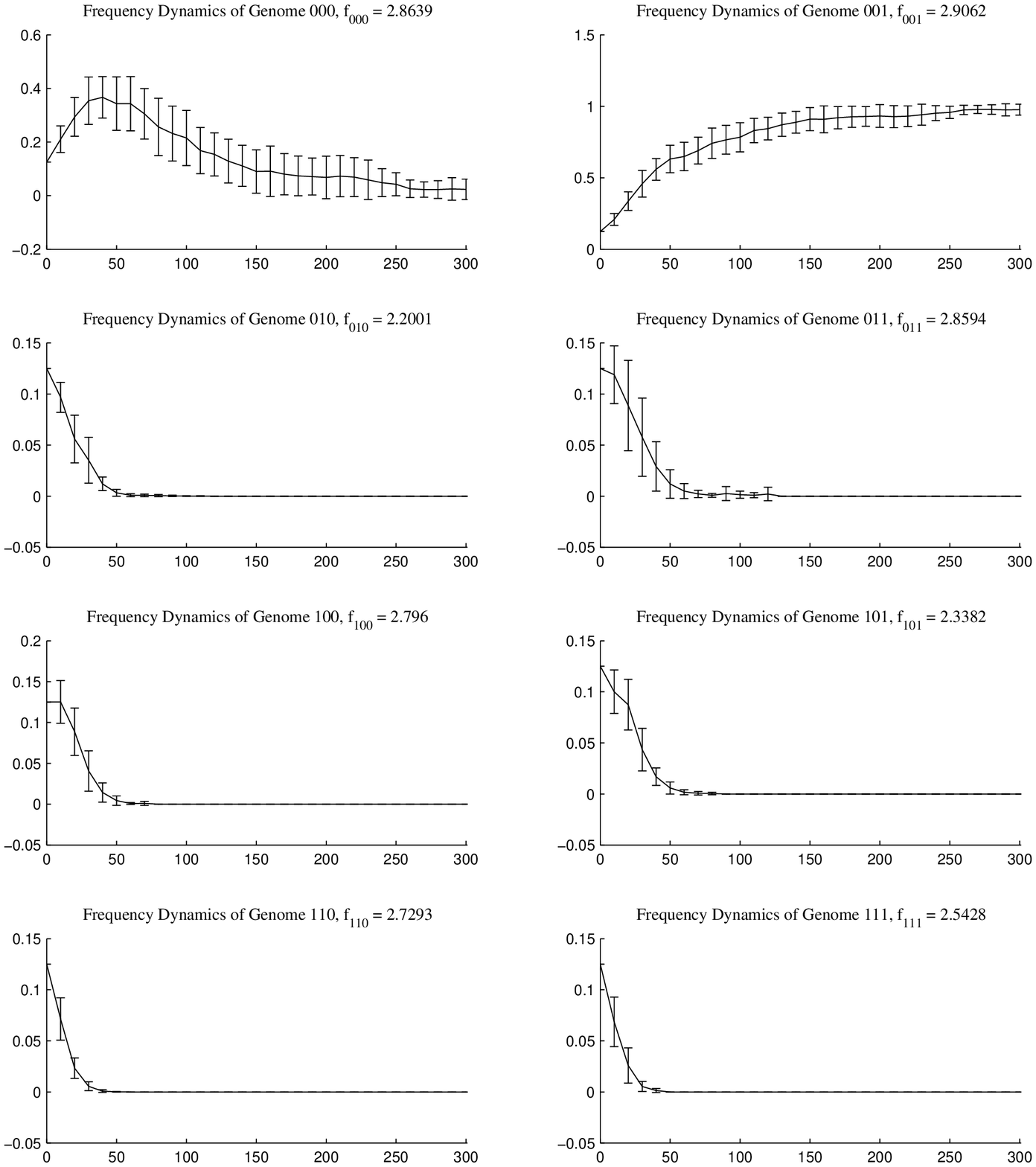}
\caption{Results of Experiment 7 ($o=3$, $N=1000$, $r=10$). A series of eight  plots showing the dynamics of the eight genomes in $\mathfrak B_3$ under the action of \textsc{sfsga}. The independent axis in each plot shows the generation number, and the dependent axis gives the average frequency of  a genome in a population averaged over 10 trials. The title of each plot displays the genome and its f-value. The thin black line plots the average frequency dynamics of the genome under the action of \textsc{sfsga} at every tenth generation. The error bars show one standard deviation above and below the average. The average f-value of the genomes in this experiment is $2.6545$.}\label{rescuingexp1}
\end{figure}

\begin{figure}
\includegraphics[height=16cm, width=\textwidth]{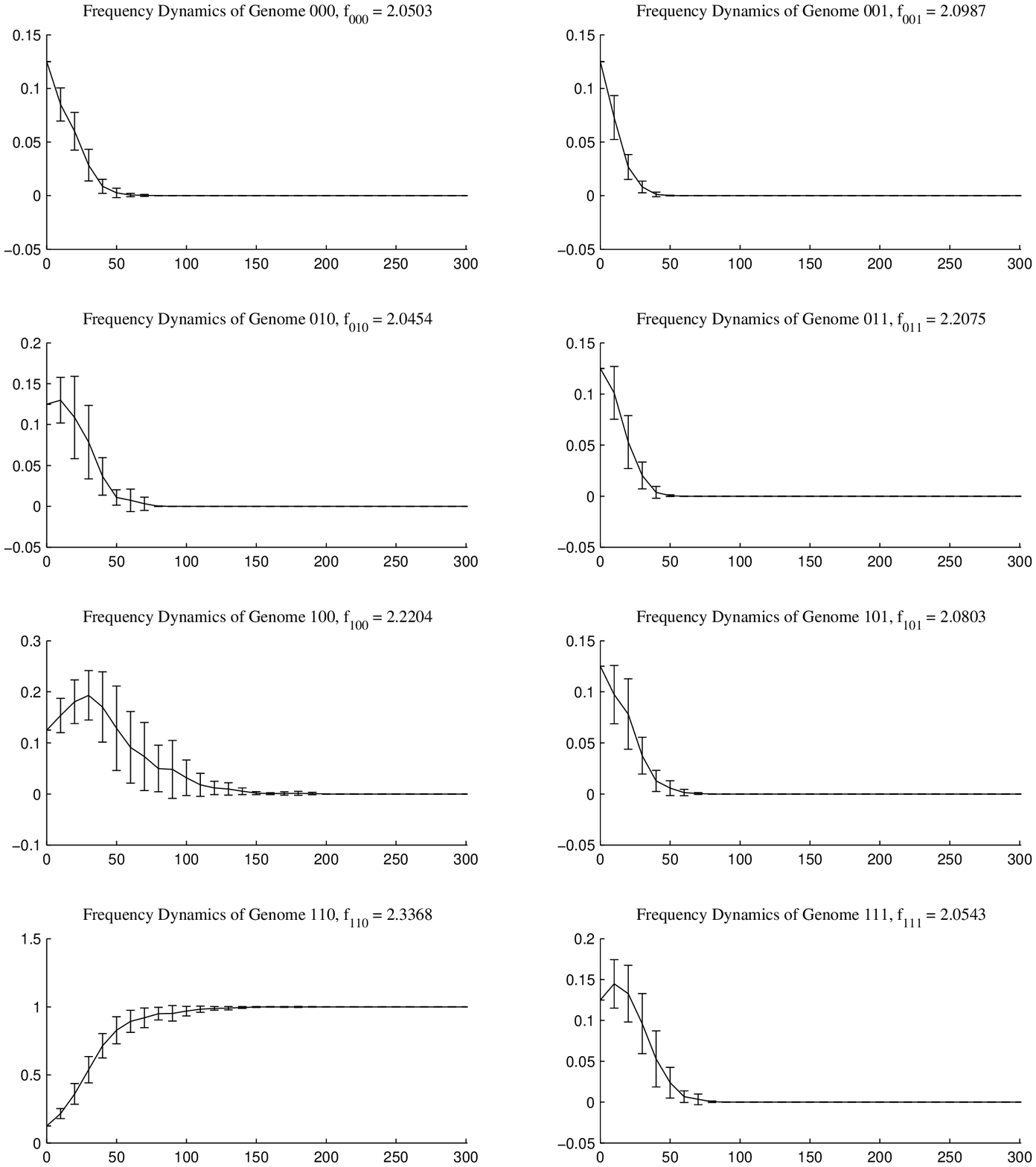}
\caption{Results of Experiment 8 ($o=3$, $N=1000$, $r=10$). A series of eight  plots showing the dynamics of the eight genomes in $\mathfrak B_3$ under the action of \textsc{sfsga}. The independent axis in each plot shows the generation number, and the dependent axis gives the average frequency of  a genome in a population averaged over 10 trials. The title of each plot displays the genome and its f-value. The thin black line plots the average frequency dynamics of the genome under the action of \textsc{sfsga} at every tenth generation. The error bars show one standard deviation above and below the average. The average f-value of the genomes in this experiment is  $2.1367$.}\label{rescuingexp2}
\end{figure}

\begin{figure}
\includegraphics[height=16cm, width=\textwidth]{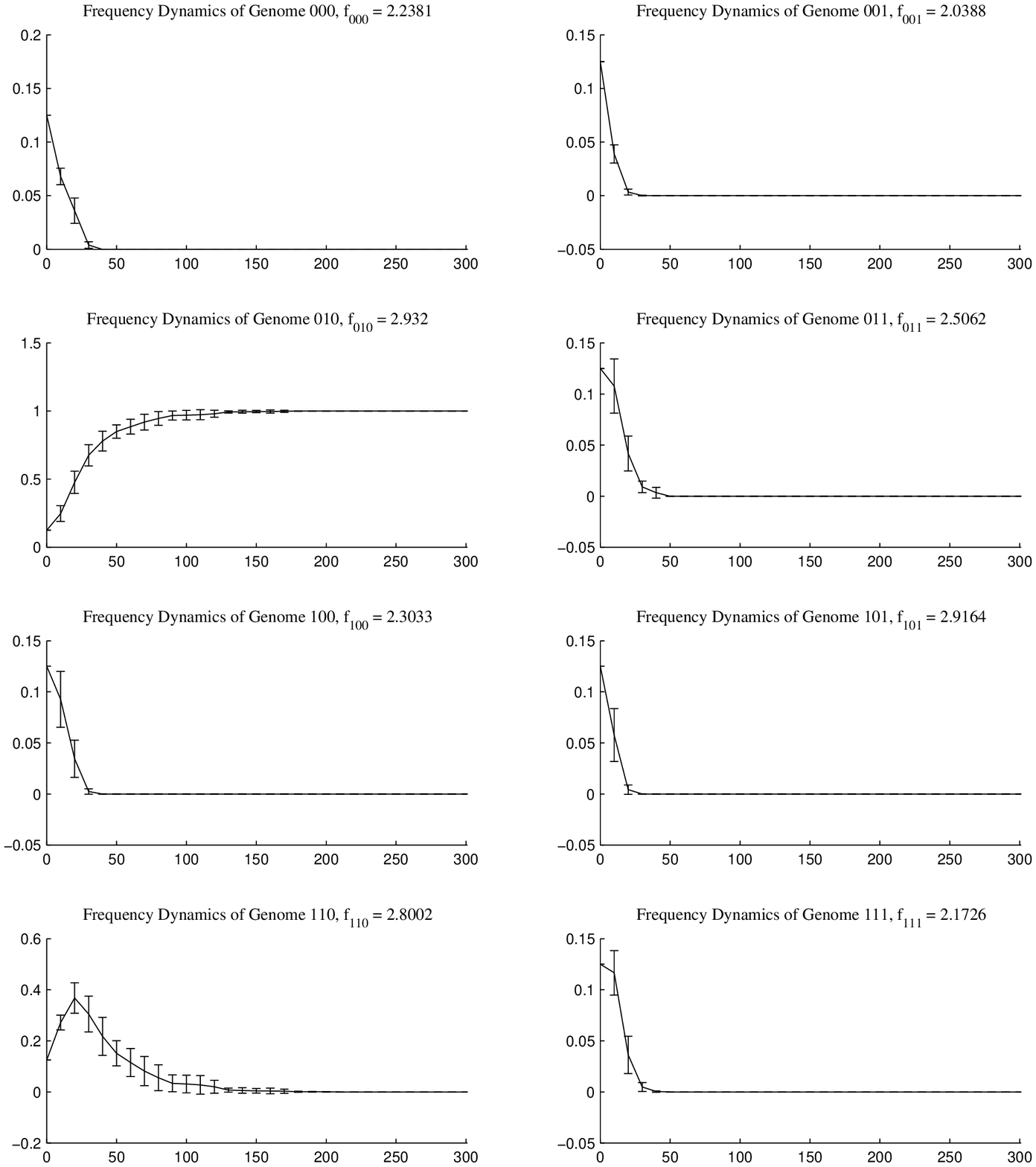}
\caption{Results of Experiment 9 ($o=3$, $N=1000$, $r=10$). A series of eight  plots showing the dynamics of the eight genomes in $\mathfrak B_3$ under the action of \textsc{sfsga}. The independent axis in each plot shows the generation number, and the dependent axis gives the average frequency of  a genome in a population averaged over 10 trials. The title of each plot displays the genome and its f-value. The thin black line plots the average frequency dynamics of the genome under the action of \textsc{sfsga} at every tenth generation. The error bars show one standard deviation above and below the average. The average f-value of the genomes in this experiment is  $2.4885$.}\label{rescuingexp3}
\end{figure}

\begin{figure}
\includegraphics[height=16cm, width=\textwidth]{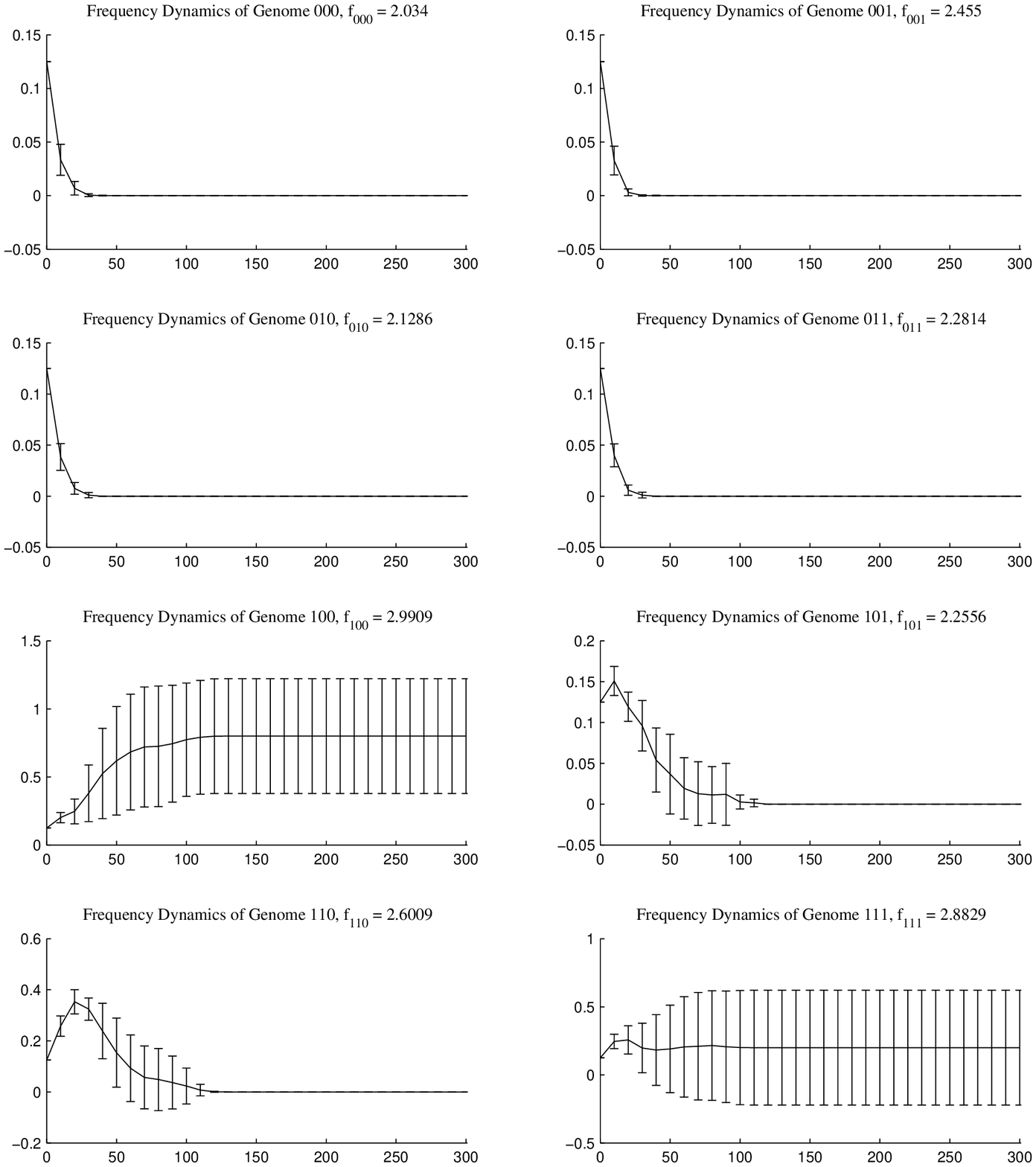}
\caption{Results of Experiment 10 ($o=3$, $N=1000$, $r=10$). A series of eight  plots showing the dynamics of the eight genomes in $\mathfrak B_3$ under the action of \textsc{sfsga}. The independent axis in each plot shows the generation number, and the dependent axis gives the average frequency of  a genome in a population averaged over 10 trials. The title of each plot displays the genome and its f-value. The thin black line plots the average frequency dynamics of the genome under the action of \textsc{sfsga} at every tenth generation. The error bars show one standard deviation above and below the average. The average f-value of the genomes in this experiment is  $2.4537$.}\label{rescuingexp4}
\end{figure}

\begin{figure}
\includegraphics[height=16cm, width=\textwidth]{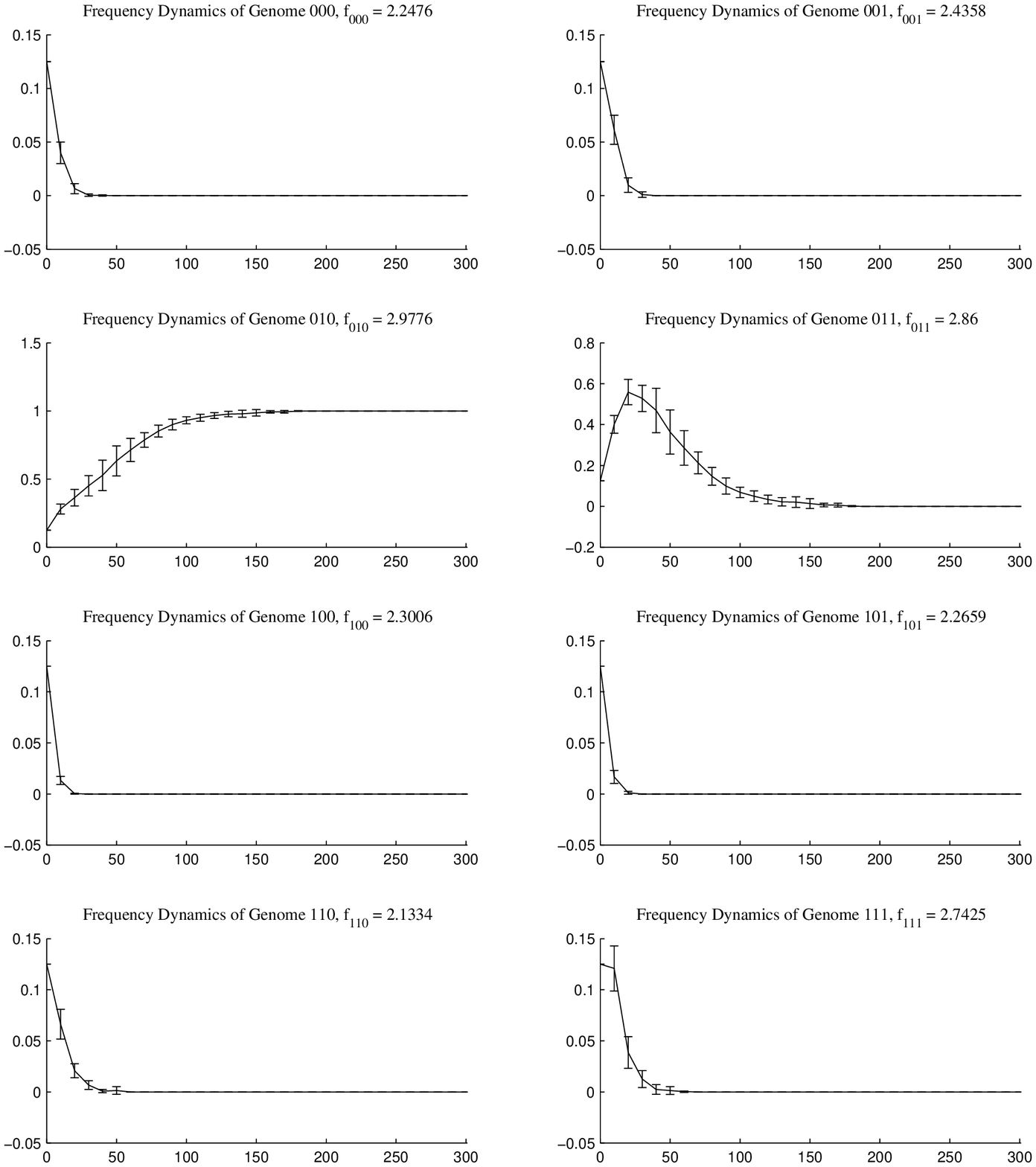}
\caption{Results of Experiment 11 ($o=3$, $N=1000$, $r=10$). A series of eight  plots showing the dynamics of the eight genomes in $\mathfrak B_3$ under the action of \textsc{sfsga}. The independent axis in each plot shows the generation number, and the dependent axis gives the average frequency of  a genome in a population averaged over 10 trials. The title of each plot displays the genome and its f-value. The thin black line plots the average frequency dynamics of the genome under the action of \textsc{sfsga} at every tenth generation. The error bars show one standard deviation above and below the average. The average f-value of the genomes in this experiment is  $2.3625$.}\label{rescuingexp5}
\end{figure}

\begin{figure}
\includegraphics[height=16cm, width=\textwidth]{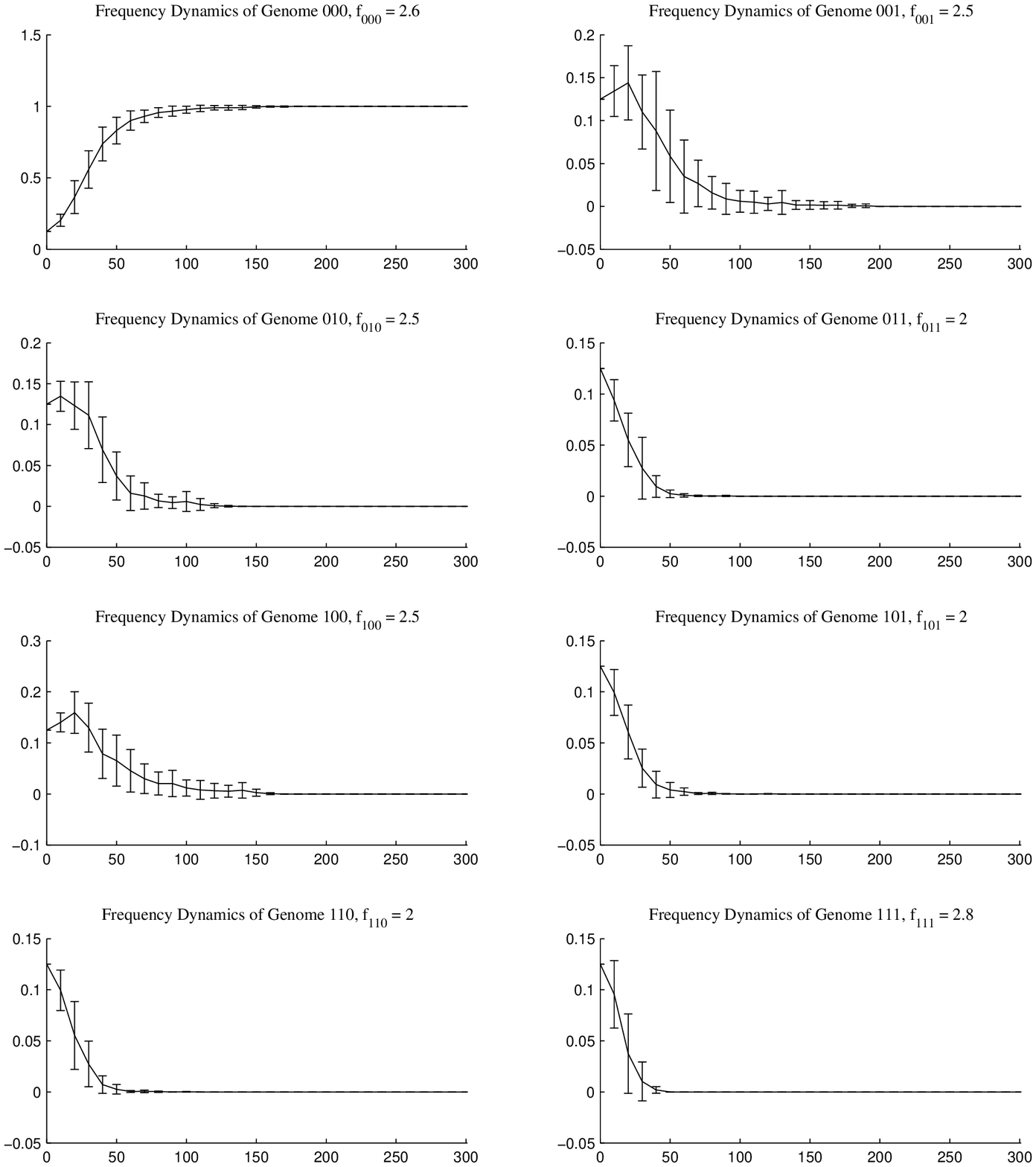}
\caption{Results of Experiment 12 ($o=3$, $N=1000$, $r=10$). A series of eight  plots showing the dynamics of the eight genomes in $\mathfrak B_3$ under the action of \textsc{sfsga}. The independent axis in each plot shows the generation number, and the dependent axis gives the average frequency of  a genome in a population averaged over 10 trials. The title of each plot displays the genome and its f-value. The thin black line plots the average frequency dynamics of the genome under the action of \textsc{sfsga} at every tenth generation. The error bars show one standard deviation above and below the average. The average f-value of the genomes in this experiment is  $2.6545$.}\label{rescuingexp6}
\end{figure}

Figures [\ref{rescuingexp1}--\ref{rescuingexp6}] present the results of a series of six experiments in which we plot the frequency dynamics of the genomes in $\mathfrak B_3$ under the action of \textsc{sfsga} (with uniform crossover, fitness proportional selection and population size $1000$). Each experiment uses a different assignment of f-values to the genomes in $\mathfrak B_3$. In all experiments except the last one, the f-values are  randomly chosen from the interval [2,3] and assigned to the genomes in $\mathfrak B_o$. From the results it can be seen that in each case the genome with the highest frequency at the end of 300 generations is one with an above average f-value. However it should be noted that this genome isn't always the one with the highest f-value (see experiments 10 and 12). From these results it can be concluded that  \textsc{sga} as described in the preceding paragraph \emph{can} effect an increase in the frequency  of a low-order schema with above average fitness \emph{even when the defining bits of that schema are widely dispersed} (in the example we are considering, the defining bits are almost a billion loci apart!)

Like many unfounded assumptions about the SGA, the assumption that it is incapable of performing the feat described above originated in Holland's seminal work \citep{holland75:_adapt_natur_artif_system}, and subsequently received unqualified support in the popular textbook by \citet{Goldberg:1989:GAS}. Overcoming this perceived limitation is the raison d'etre for at least two new adaptation algorithms --- messy GA \citep{goldberg:1989:mgamafr,desInnov} and LLGA \citep{harik97learning,desInnov}. Furthermore the inventors of several other adaptation algorithms --- CGA \citep{journals/tec/HarikLG99}, ECGA \citep{harik99linkage}, FDA \citep{muehlenbein:1999:fdaadf}, LFDA \citep{muehlenbein:2001:earsd}, BOA \citep{Pelikan:98aa,desInnov}, hBOA \citep{pelikan:2001:HBOABOANLS}, SEAM \citep{oai:eprints.ecs.soton.ac.uk:12006,watsonBook}, etc. --- have touted the superiority of their algorithms vis-a-vis the SGA on this matter. The simple experiments in this section show that on this matter, the SGA is more powerful than it is commonly thought to be.

\section{Conclusion}
The biosphere is replete with organisms that are exquisitely well adapted to the environmental niches they inhabit. Natural sexual evolution has been crucial to the generation of what are arguably the most highly adapted of these organisms --- cheetahs, owls, humans etc. A deeply intriguing idea is that we can build adaptation algorithms which, at an abstract level, mimic the behavior of natural sexual evolution, and in doing so, ``harness'' something of the adaptive power of this incredibly effective process.

But what \emph{is} the abstract level at which natural sexual evolution should be mimicked? In other words given everything we know about natural sexual evolutionary systems, how do we distinguish between aspects of these systems which are essential to their adaptive power, and those which can be viewed as ``mere biological detail" and need not be simulated, especially when taking a first swipe at building an artificial evolutionary systems which harnesses the power of natural sexual evolution? For instance is it necessary for such ``first-order" artificial evolutionary systems to simulate hydrogen bonding between the bases of  DNA strands? What about diploidy, or the way genomes of organisms are comprised of multiple chromosomes? And how crucial is the fact that crossover takes place between homologous chromosomes during meiosis? Engineers of yesteryear faced a similar quandary when trying to ascertain just what it is about birds that gives them their capacity for flight? Stories of inventors in feathered suits jumping to their deaths off cliffs and buildings bear testament to the fact that that our initial answers to such questions are often incorrect. It was only after the realization that birds were ``using" Bernoulli's principle to stay aloft that researchers began to make any real progress towards building machines that successfully harnessed the principle underlying birds' capacity for soaring. One can infer the following general rule from this example: without a good understanding of exactly why a natural system exhibits a certain useful phenomenon, efforts to build artificial systems that exhibit the same phenomenon by mimicking the natural system will be misguided and are unlikely to be successful.

The field of Population Genetics stems from the efforts of its founders --- Fisher, Wright and Haldane --- to  reconcile Darwin's theory of adaptation by natural selection with Mendel's theory of  genetics (odd as it may now seem, these two theories were once thought to contradict each other. See \citealp{stanfordPopGenetics}). The literature of this field seems like the most reasonable place to look for answers to questions about how and why adaptation occurs in natural sexually reproducing populations. Unfortunately Population Genetics does not hold ready answers to these questions. Indeed the differing theories of Fisher and Wright regarding exactly this issue has been the subject of a longstanding and ongoing debate \citep{WadeGoodnight, brodie2000}. Significant empirical evidence has been gathered by both sides in support of their respective positions yet a definitive answer has not emerged. The absence of a definitive answer makes it difficult to make principled decisions about the level of detail which must be present in an artificial system which seeks, through mimicry, to harness something of the adaptive power of natural sexual evolution.

Let us return to the analogy with the field of aviation that we introduced above. For the sake of argument let us suppose that in the age before the discovery of Bernoulli's principle someone had, for some reason or the other, succeeded in inventing a simple winged machine which a) mimicked birds at some relatively abstract level, and b) was capable of soaring long distances (even if  slowly, or inefficiently). Such a machine would immediately be incredibly interesting because when compared to the complex body of a bird, such a machine would be much more amenable to analysis. The principle underlying this machine's ability to soar, once derived, could then be used to build ``better" soaring machines. This underlying principle would also play a very important part in the development of an accurate theory of why birds can soar\footnote{If the inventor of the winged machine gives an incorrect reason for why his machine can soar, that would probably slow down the progression described above, but it would take away nothing from the importance of the winged machine itself.}. The implications of this vignette for the importance of the SGA to the fields of Evolutionary Computation and Evolutionary Biology should be evident. The SGA should thus be viewed as a `lucky break', one that can and should be exploited for its potential to advance theories and applications of the adaptive capacity of sexual evolution.

Let us spell out the importance of studying the SGA's capacity for adaptation. As models of sexual evolutionary systems go, the SGA is arguably the simplest one which has regularly been observed to adapt high-quality solutions despite the almost certain presence of non-trivial epistasis between genomic loci, in other words, despite its application to problems whose representations are in all likelihood riddled with local optima.  Because of its effectiveness in spite of its simplicity, the SGA is a model of sexual evolution that is highly likely to a) yield an explanation for the incredible adaptive capacity of sexual evolution and b) precipitate the identification of classes of non-trivial epistasis which do not pose much of a problem for sexual evolution. The SGA is of course not the last word in evolution inspired adaptive systems. Efforts to extend this algorithm in ways that  ``increase its adaptive power" should and have been made. However if, while attempting to extend the SGA, one works within a flawed paradigm, one is unlikely to capitalize on, and may even compromise, whatever ``power" the SGA derives, by virtue of imitation,  from natural sexual evolution. A non-dogmatic study of how SGAs perform adaptation has the potential to yield a theory which accurately explains the reasons behind the SGA's frequent success. Such a theory will probably usher in a new paradigm within which fruitful research into the construction of more ``powerful" evolutionary algorithms can proceed. If such a theory differs significantly from those of Fisher and Wright it is likely to have deep implications for the field of Population Genetics and the larger field of Evolutionary Biology (within which several basic questions --- why sex? Why punctuated equilibrium? Why diploidy and polyploidy?  Why speciation? What is the unit of selection? --- have yet to receive satisfactory answers). Finally, a study that reveals the SGA's capacity for adaptation will also undoubtedly reveal classes of problems that SGAs can efficiently solve. These classes of problems may prove useful to machine learning researchers in their efforts to find semi-principled reductions of difficult learning problems to problems for which robust and efficient solvers exist.

This paper makes two concrete contributions. Firstly, in sections \ref{MathematicalPreliminaries}--\ref{expval} we have derived results which we believe are relevant to the riddle of the SGA's capacity for adaptation. Secondly in section \ref{rescuing} we have presented results which show that an oft perceived shortcoming of the SGA is misplaced. The following two paragraphs expand upon these contributions.

As we discussed in section \ref{TowPrincTheory}, a promising way to understand the effect of selection and variation on the composition of the evolving population of an SGA is by understanding the multi-generational effect of these operations on the search distribution of the SGA. One way to study an evolving high-dimensional distribution is to study its evolving multivariate marginals. In this paper we derive conditions under which a multivariate marginal of the search distribution of an SGA, with an infinite population of long genomes, can be approximated over multiple generations. In other words we derive conditions under which the frequency dynamics of some family of schemata under the action of an SGA, with an infinite population of long genomes, can be approximated over multiple generations. The conditions we derived in this paper are much weaker than those derived by \citet{conf/gecco/WrightVR03}. This makes our result more useful. The conclusions reached in section \ref{suffConds} are reached by making small leaps of intuition. In section \ref{expval} we experimentally validated these conclusions. Our validation, though  indirect, is based on assumptions and modeling decisions which are, in our opinion, uncontroversial.

Besides being incorrectly used to support an outlandish hypothesis about what the SGA \emph{can} do (hierarchical building block assembly), Holland's Schema Theorem has also heavily shaped opinion about what the SGA \emph{cannot} do.  Following the experiments by \citet{mitchell:1992:rrgaflgp}, and \citet{forrest93relative}, the perceived abilities of the SGA stand compromised, yet the perceived limitations of the SGA have remained unchanged. The SGA is currently thought to be incapable of increasing the frequency of a low-order schema with higher than average fitness when the defining length of that schema is large \citep{goldberg:1989:mgamafr, desInnov}.  In section \ref{rescuing} we argued that this perception is just plain wrong --- an SGA \emph{can} increase the frequency of a low-order schema with higher than average fitness even when the defining length of that schema is large.

In closing we briefly mention that we have recently obtained a new,  simple, and (in our opinion) satisfying theory which explains the SGA's remarkable capacity for adaptation. We have also identified a class of hard statistical problems such that a) the problems in this class can be solved efficiently and robustly by an SGA, and b) this class is likely to be a useful target of machine learning reductions. All of this will soon appear in our forthcoming dissertation. Our theory relies crucially on the SGA's ability to increase the frequency of schemata of low order and above average fitness, even when the defining bits of those schemata are widely dispersed.

\bibliographystyle{plainnat}
\bibliography{c:/mystuff/mycreations/0Work/refs}

\appendix \large \vspace{0.3cm}\noindent \textbf{Appendix}\small

\ifthenelse{\boolean{showProofs}}{

\begin{lemma}
For any finite set $X$, and any metric space $(\Upsilon,d)$,let  $\mathcal A:\Upsilon\rightarrow \Lambda^X$ and let $\mathcal B:X\rightarrow [\Upsilon\rightarrow [0,1]]$ be functions\footnote{For any sets $X, Y$ we use the notation $[X\rightarrow Y]$ to denote the set of all functions from $X$ to $Y$} such that for any $h\in \Upsilon$, and any $x\in X$, $(\mathcal B(x))(h)=(\mathcal A(h))(x)$. For any $x\in X$, and for any $h^*\in \Upsilon$, if the following statement is true
\begin{align*} \forall x\in X, \forall \epsilon_x>0, \exists \delta_x>0, \forall h\in \Upsilon, d(h,h^*)<\delta_x\Rightarrow |(\mathcal B(x))(h)-(\mathcal B(x))(h^*)|<\epsilon_x\end{align*}Then we have that \begin{align*}\forall \epsilon>0, \exists
\delta>0,  \forall h\in \Upsilon, d(h, h^*)<\delta \Rightarrow d(\mathcal A(h),\mathcal A(h^*))<\epsilon\end{align*}
\end{lemma}
This lemma says that $\mathcal A$ is continuous at $h^*$ if for all $x\in X$, $\mathcal B(x)$ is continuous at $h^*$.\\
\textsc{Proof:  } We first prove the following two claims

\begin{clam}
\begin{multline*}\forall x\in X \textrm{ s.t. } (\mathcal B(x))(h^*)>0,  \forall \epsilon_x>0, \exists \delta_x>0, \forall h\in \Upsilon,\\
    d(h,h^*)<\delta_x\Rightarrow |(\mathcal B(x))(h)-(\mathcal B(x))(h^*)|<\epsilon_x.(\mathcal B(x))(h^*)\end{multline*}
\end{clam}
This claim follows from the continuity of $\mathcal B(x)$ at $h^*$ for all $x\in X$ and the fact that $(\mathcal B(x))(h^*)$ is a positive constant w.r.t. $h$.
\begin{clam} For all $h\in  \Upsilon$
\[\sum_{\substack{x\in X \text{s.t.}\\(\mathcal A(h^*))(x)>\\(\mathcal A(h))(x)}}|(\mathcal A(h^*))(x)-(\mathcal A(h))(x)|=
 \sum_{\substack{x\in X \text{s.t.}\\(\mathcal A(h))(x)>\\(\mathcal A(h^*))(x)}}|(\mathcal A(h))(x)-(\mathcal A(h^*))(x)|\]
 \end{clam}
\noindent The proof of this claim is as follows: for all $h\in \Upsilon$,
{\allowdisplaybreaks\begin{align*}
&\sum_{x\in X}(\mathcal A(h^*)(x))-(\mathcal A(h))(x)=0\\
&\Rightarrow \sum_{\substack{x\in X \text{s.t.}\\(\mathcal A(h^*))(x)>\\(\mathcal A(h))(x)}}(\mathcal A(h^*))(x)-(\mathcal A(h))(x) -
 \sum_{\substack{x\in X \text{s.t.}\\(\mathcal A(h))(x)>\\(\mathcal A(h^*))(x)}}(\mathcal A(h))(x)-(\mathcal A(h^*))(x)=0\\
&\Rightarrow \sum_{\substack{x\in X \text{s.t.}\\(\mathcal A(h^*))(x)>\\(\mathcal A(h))(x)}}(\mathcal A(h^*))(x)-(\mathcal A(h))(x)=
 \sum_{\substack{x\in X \text{s.t.}\\(\mathcal A(h))(x)>\\(\mathcal A(h^*))(x)}}(\mathcal A(h))(x)-(\mathcal A(h^*))(x)\\
 &\Rightarrow \bigg|\sum_{\substack{x\in X \text{s.t.}\\(\mathcal A(h^*))(x)>\\(\mathcal A(h))(x)}}(\mathcal A(h^*))(x)-(\mathcal A(h))(x)\Bigg|=\Bigg|
 \sum_{\substack{x\in X \text{s.t.}\\(\mathcal A(h))(x)>\\(\mathcal A(h^*))(x)}}(\mathcal A(h))(x)-(\mathcal A(h^*))(x)\Bigg|\\
 &\Rightarrow \sum_{\substack{x\in X \text{s.t.}\\(\mathcal A(h^*))(x)>\\(\mathcal A(h))(x)}}|(\mathcal A(h^*))(x)-(\mathcal A(h))(x)|=
 \sum_{\substack{x\in X \text{s.t.}\\(\mathcal A(h))(x)>\\(\mathcal A(h^*))(x)}}|(\mathcal A(h))(x)-(\mathcal A(h^*))(x)|
\end{align*}
\noindent We now prove the lemma. Using claim 1 and the fact that $X$ is finite, we get that $\forall \epsilon>0$, $\exists \delta>0$, $\forall h\in[X\rightarrow\mathbb R]$ such that $d(h,h^*)<\delta$,
\begin{align*}
&\sum_{\substack{x\in X \text{s.t.}\\(\mathcal A(h^*))(x)>\\(\mathcal A(h))(x)}}|(\mathcal B(x))(h^*)-(\mathcal B(x))(h)|<\sum_{\substack{x\in X \text{s.t.}\\(\mathcal A(h^*))(x)>\\(\mathcal A(h))(x)}}\frac{\epsilon}{2}.(\mathcal B(x))(h^*)\\
&\Rightarrow \sum_{\substack{x\in X \text{s.t.}\\(\mathcal A(h^*))(x)>\\(\mathcal A(h))(x)}}|(\mathcal A(h^*))(x)-(\mathcal A(h))(x)|<\sum_{\substack{x\in X \text{s.t.}\\(\mathcal A(h^*))(x)>\\(\mathcal A(h))(x)}}\frac{\epsilon}{2}.(\mathcal A(h^*))(x)\\
&\Rightarrow \sum_{\substack{x\in X \text{s.t.}\\(\mathcal A(h^*))(x)>\\(\mathcal A(h))(x)}}|(\mathcal A(h^*))(x)-(\mathcal A(h))(x)|<\frac{\epsilon}{2}\quad\quad \qed
\end{align*}
\noindent By Claim 2 and the result above, we have that $\forall \epsilon>0$, $\exists \delta>0$, $\forall h\in[X\rightarrow\mathbb R]$ such that $d(h,h^*)<\delta$,
\begin{align*}
  \sum_{\substack{x\in X \text{s.t.}\\(\mathcal A(h))(x)>\\(\mathcal A(h^*))(x)}}|(\mathcal A(h))(x)-(\mathcal A(h^*))(x)|<\frac{\epsilon}{2}
\end{align*}
\noindent Therefore, given the two previous results, we have that $\forall \epsilon>0$, $\exists \delta>0$, $\forall h\in[X\rightarrow\mathbb R]$ such that $d(h,h^*)<\delta$,
\begin{align*}
&\sum_{x\in X}|(\mathcal A(h))(x)-(\mathcal A(h^*)(x))|<\epsilon \quad\quad \qed
\end{align*}}
\begin{lemma} \label{lemmaB}
Let $X$ be a finite set, and let $T\in \Lambda^X_m$ be a transmission function. Then for any $p'\in \Lambda^X$ and any $\epsilon>0$, there exists a $\delta>0$ such that for any $p\in \Lambda^X$,
\[d(p\,,\,p')<\delta\Rightarrow d(\mathcal V_Tp\,,\, \mathcal V_Tp')<\epsilon\]
\end{lemma}
\noindent \textit{Sketch of Proof:  } Let $\mathcal A:\Lambda^X\rightarrow\Lambda^X$ be defined such that $(A(p))(x)=(\mathcal V_Tp)(x)$. Let $\mathcal B:X\rightarrow[\Lambda^X\rightarrow [0,1]]$ be defined such that $(\mathcal B(x))(p)=(\mathcal V_Tp)(x)$. The reader can check that for any $x\in X$, $\mathcal B(x)$ is a continuous function. The application of lemma 1 completes the proof. \\

\noindent By similar arguments, we obtain the following two lemmas.
\begin{lemma} \label{lemmaC}
  Let $X$ be a finite set, and let $f:X\rightarrow \mathbb R^+$ be a function. Then for any $p'\in \Lambda^X$ and any $\epsilon>0$, there exists a $\delta>0$ such that for any $p\in \Lambda^X$,
  \[d(p\,,\,p')<\delta\Rightarrow d(\mathcal S_fp\,,\, \mathcal S_fp')<\epsilon\]
\end{lemma}

\begin{lemma} \label{lemmaD}
  Let $X$ be a finite set, and let $p\in \Lambda^X$ be a distribution. Then for any $f'\in [X\rightarrow \mathbb R^+]$, and any $\epsilon>0$, there exists a $\delta>0$ such that for any $f\in[X\rightarrow \mathbb R^+]$,
  \[d(f\,,\,f')<\delta\Rightarrow d(\mathcal S_fp\,,\, \mathcal S_{f'}p)<\epsilon\]
\end{lemma}}{}
\end{document}